\documentclass[%
 reprint,
 amsmath,amssymb,
 aps, 
 pre,
 showkeys
]{revtex4-1}

\usepackage{graphicx}
\usepackage{dcolumn}
\usepackage{bm}

\usepackage{my-style}
\graphicspath{{figs/}}

\allowdisplaybreaks[4]

\begin{document}

\preprint{APS/123-QED}

\title{Statistical Analysis of Loopy Belief Propagation in Random Fields}

\author{Muneki Yasuda}  
\affiliation{%
 Graduate School of Science and Engineering, Yamagata University, Japan.\\
 CREST, JST (Yamagata University)
}%
\author{Shun Kataoka and Kazuyuki Tanaka}%
\affiliation{%
 Graduate School of Information Sciences, Tohoku University, Japan.\\
 CREST, JST (Tohoku University)
}%


\begin{abstract}
Loopy belief propagation (LBP), which is equivalent to the Bethe approximation in statistical mechanics, is a message-passing-type inference method that is widely used to analyze systems based on Markov random fields (MRFs).
In this paper, we propose a message-passing-type method to analytically evaluate the quenched average of LBP in random fields by using the replica cluster variation method. 
The proposed analytical method is applicable to general pair-wise MRFs with random fields whose distributions differ from each other 
and can give the quenched averages of the Bethe free energies over random fields, which are consistent with numerical results. 
The order of its computational cost is equivalent to that of standard LBP. 
In the latter part of this paper, we describe the application of the proposed method to Bayesian image restoration, in which we observed that our theoretical results are in good agreement with the numerical results for natural images.
\end{abstract}

\pacs{Valid PACS appear here}
\keywords{random field model, loopy belief propagation, replica method, cluster variation method, Bayesian image restoration}
\maketitle


\section{Introduction} \label{sec:Introduction}

Loopy belief propagation (LBP)~\cite{Pearl1988}, which is a message-passing-type inference method, 
is widely prevalent in various fields, including computer science, as a powerful tool for statistical procedures in systems based on Markov random fields (MRFs)~\cite{Opper&Saad2001,Mezard&Montanari2009}. 
LBP is equivalent to the Bethe approximation in statistical mechanics~\cite{Kaba&Saad1998,GLBP2005} 
and is also known as the cavity method. 
An analysis of the statistical behaviors of LBP is important to develop an understanding of LBP.
In this paper, we focus on LBP in pair-wise MRFs with random fields and we present a statistical analysis of it, namely, an analysis of the quenched average of LBP over random fields.  
The topic of pair-wise MRFs in random fields is an important research field in statistical mechanics~\cite{RFIM1977,RFIM2010}. 
As described below, a statistical analysis of LBP in random fields is also important for the field of Bayesian signal processing in computer science.


Bayesian image restoration~\cite{Geman&Geman1984}, in which images degraded by noise are restored using the Bayesian framework, 
is an important generic technique for various types of signal processing. 
Suppose that there is an original image and that the original image is degraded through a specific noise process.
We observe only the degraded image as the input, and we want to produce the restored image as the output. 
From the statistical mechanics point of view, the standard framework of Bayesian image restoration 
corresponds to the framework of a two-dimensional ferromagnetic spin model in random fields~\cite{Nishimori2001,TanakaReview2002}. 
In this correspondence, the input image, namely, the degraded image, is regarded as the random fields in the Bayesian image restoration system. 

Since the model used in the Bayesian image restoration system is designed by using an intractable pair-wise MRF, LBP is often applied to implement it.
Hence, in the evaluation of the statistical performance of the implemented image restoration system, 
we encounter the evaluation of the quenched average of LBP over the random fields, namely, over the input images. 
For this purpose, for Ising systems, the authors proposed an analytical evaluation method for it~\cite{KYT2010,TKY2010}. 
In the previous method, the evaluation of the quenched average of LBP is reduced to solving simultaneous integral equations with respect to the distributions of the messages.
However, the method is not very practical, because the computational cost of solving the integral equations is considerable 
and its approximation accuracy is poor. 
Furthermore, the method cannot evaluate the quenched average of the free energy and is formulated only in Ising systems.

In this paper, we propose a new analytical method for evaluating the quenched average of LBP over random fields 
based on the idea of the replica cluster variation method (RCVM)~\cite{RCVM2010, RCVM2013}. 
The presented method allows the quenched average of the Bethe free energies over random fields in general pair-wise MRFs to be evaluated, unlike the previous method.

The remaining part of this paper is organized as follows.
A brief explanation of LBP is given in section \ref{sec:LBP}.
Section \ref{sec:ProposeMethod} constitutes the main part of the paper.
The proposed method is shown in section \ref{sec:ReplicaMessagePassing}, 
and some numerical results for checking its validity are shown in section \ref{sec:check-validity}. 
In section \ref{sec:exact-case}, we show a case that is exactly solvable by the present method. 
In section \ref{sec:ImageRestoration}, we explain the framework of the framework of Bayesian image restoration and 
compute the statistical performance of the Bayesian image restoration system using the proposed method.
Finally, section \ref{sec:conclusion} closes the paper with concluding remarks.

\section{Loopy Belief Propagation in Random Fields}\label{sec:LBP}

\subsection{Model Definition}

Let us consider an undirected graph $G(V,E)$ consisting of $n$ vertices and some edges, where $V=\{1,2,\ldots,n\}$ is the set of vertices, and 
$E = \{\{i,j\}\}$ is the set of edges between a pair of vertices, where $\{i,j\}$ denotes the undirected edge between vertices $i$ and $j$. 
On the graph, with the discrete random variables $\bm{S} \in \{S_i \mid i \in V\}$, let us define the pair-wise MRF expressed by
\begin{align}
P(\bm{S}\mid \bm{h},\beta) &:= \frac{1}{Z(\bm{h}, \beta)}\exp  \big( -\beta H(\bm{S}; \bm{h})\big),
\label{eq:MRF}
\end{align}
where 
\begin{align*}
H(\bm{S}; \bm{h}):=-\sum_{i \in V} \phi_i(S_i, h_i) - \sum_{\{i,j\} \in E}\psi_{i,j}(S_i, S_j)
\end{align*}
is the Hamiltonian of the MRF.
Here, $\phi_i(S_i, h_i)$ is a specific function of the variable $S_i$ and the random field $h_i$ on vertex $i$,  
and $\phi_{i,j}(S_i, S_j)$ is a specific function on edge $\{i,j\}$.
The notations, $Z$ and $\beta$, are the partition function and inverse temperature, which takes a positive value, respectively. 
In this paper, only random fields $\bm{h}$ are treated as the quenched parameters.

\subsection{Loopy Belief Propagation}
\label{sec:LBP-detail}

\begin{figure}[htb]
\begin{center}
\includegraphics[height=2.5cm]{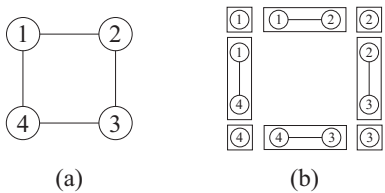}
\end{center}
\caption{(a) Square graph with four vertices. (b) Cluster decomposition of the Bethe approximation of the CVM for (a). The vertices and two connected vertices are selected as the clusters.}
\label{fig:Bethe-decomp}
\end{figure}

It is known that LBP is derived from the minimum condition of the variational Bethe free energy~\cite{GLBP2005}.
In this section, we give a brief explanation of the derivation of LBP according to the cluster variation method (CVM)~\cite{CVM1951,CVM-review2005}.
The free energy of the MRF in equation (\ref{eq:MRF}) is defined by
\begin{align}
F(\bm{h}, \beta)&:= \sum_{\bm{S}}H(\bm{S}; \bm{h})P(\bm{S}\mid \bm{h}, \beta) \nn
\aldef + \frac{1}{\beta}\sum_{\bm{S}}P(\bm{S}\mid \bm{h}, \beta) \ln P(\bm{S}\mid \bm{h}, \beta).
\label{eq:FreeEnergy}
\end{align}
In the Bethe approximation in the CVM, we approximate the MRF by
\begin{align}
P(\bm{S}\mid \bm{h},\beta)\approx  \frac{\big(\prod_{i \in V} b_i(S_i)\big)\big(\prod_{\{i,j\} \in E}b_{i,j}(S_i,S_j)\big)}{\prod_{\{i,j\} \in E}b_i(S_i)b_j(S_j)},
\label{eq:BetheDecomp-original}
\end{align}
where $b_i(S_i)$ and $b_{i,j}(S_i,S_j)$ are the one-vertex and two-vertex marginal distributions (or the \textit{beliefs}) of the MRF. 
This approximation corresponds to the cluster decomposition shown in figure \ref{fig:Bethe-decomp}. 
The right-hand side of equation (\ref{eq:BetheDecomp-original}) is the product of the marginal distributions of the clusters 
divided by the product of the marginal distributions of the double-counted clusters.
By applying this approximation to $P(\bm{S}\mid \bm{h})$ in the logarithmic function of the last term in equation (\ref{eq:FreeEnergy}), 
we obtain the variational Bethe free energy expressed by
\begin{widetext}
\begin{align}
\mcal{F}_{\mrm{bethe}}[\{b_i, b_{i,j}\}]&:=\sum_{\bm{S}}H(\bm{S}; \bm{h})P(\bm{S}\mid \bm{h}, \beta)
+ \frac{1}{\beta}\sum_{\bm{S}}P(\bm{S}\mid \bm{h}, \beta) \ln \frac{\big(\prod_{i \in V} b_i(S_i)\big)\big(\prod_{\{i,j\} \in E}b_{i,j}(S_i,S_j)\big)}{\prod_{\{i,j\} \in E}b_i(S_i)b_j(S_j)} \nn
&\>=-\sum_{i \in V}\sum_{S_i}\phi_i(S_i, h_i) b_i(S_i) 
- \sum_{\{i,j\} \in E}\sum_{S_i, S_j}\psi_{i,j}(S_i, S_j)b_{i,j}(S_i,S_j)+\frac{1}{\beta}\sum_{i \in V}\mcal{H}_1[b_i]\nn
\aldef\,
+ \frac{1}{\beta}\sum_{\{i,j\} \in E}\big(\mcal{H}_2[b_{i,j}] - \mcal{H}_1[b_i] -\mcal{H}_1[b_j]\big),
\label{eq:BetheFreeEnergy}
\end{align}
\end{widetext}
where
\begin{align*}
\mcal{H}_1[b_i]:= \sum_{S_i}b_i(S_i) \ln b_i(S_i)
\end{align*}
and
\begin{align*}
\mcal{H}_2[b_{i,j}]:= \sum_{S_i,S_j}b_{i,j}(S_i,S_j) \ln b_{i,j}(S_i,S_j)
\end{align*}
are the one-vertex and two-vertex negative entropies. 

LBP is obtained by the variational minimization of the variational Bethe free energy with respect to the beliefs.
By minimizing the variational Bethe free energy under the normalizing constraints, $\sum_{S_i}b_i(S_i) = \sum_{S_i,S_j}b_{i,j}(S_i,S_j)=1$, 
and the marginalizing constraints, $b_i(S_i) = \sum_{S_j}b_{i,j}(S_i,S_j)$ and $b_j(S_j)=\sum_{S_i}b_{i,j}(S_i,S_j)$, 
we obtain the message-passing equation of LBP: 
\begin{align}
M_{i \to j}(S_j) &\propto \sum_{S_i}\exp \beta \big( \phi_i(S_i, h_i) + \psi_{i,j}(S_i,S_j)\big)\nn
\aleq 
\times \prod_{k \in \partial i \setminus \{j\}}M_{k \to i}(S_i), 
\label{eq:message-passing}
\end{align}
where $M_{i\to j}(S_i)$ is the \textit{message} (or the effective field) from vertex $i$ to vertex $j$. 
Using the messages satisfying the message-passing equation, we can compute the beliefs that minimize the variational Bethe free energy as
\begin{align}
b_i(S_i)&\propto \exp\big( \beta \phi_i(S_i,h_i) \big) \prod_{j \in \partial i}M_{j \to i}(S_i),
\label{eq:LBP-bi}
\end{align}
\begin{align}
&b_{i,j}(S_i,S_j) \propto  \exp \beta \big( \phi_i(S_i, h_i) + \phi_j(S_j,h_j) \nn
&+ \psi_{i,j}(S_i,S_j)\big)\prod_{k \in \partial i \setminus \{j\}}M_{k \to i}(S_i)
\prod_{l \in \partial j \setminus \{i\}}M_{l \to j}(S_j),
\label{eq:LBP-bij}
\end{align}
where $\partial i$ is the set of vertices connected to vertex $i$: $\partial i = \{ j \mid j \in V,\, \{i,j\} \in E\}$.
The Bethe free energy of the MRF is the minimum of the variational Bethe free energy, 
\begin{align*}
F_{\mrm{bethe}}(\bm{h}, \beta) := \min_{\{b_i,b_{ij}\}}\mcal{F}_{\mrm{bethe}}[\{b_i, b_{i,j}\}], 
\end{align*}
and is obtained by substituting the beliefs obtained by equations (\ref{eq:LBP-bi}) and (\ref{eq:LBP-bij}) 
into the variational Bethe free energy in equation (\ref{eq:BetheFreeEnergy}).
In LBP, the beliefs obtained by equations (\ref{eq:LBP-bi}) and (\ref{eq:LBP-bij}) are regarded as the Bethe approximations of the true marginal distributions of the MRF.
When an undirected graph $G(V,E)$, on which the MRF is defined, has no loops, 
the Bethe free energy and the beliefs are equivalent to the true free energy and the true marginal distributions of the MRF, respectively.  

The main proposal presented in this paper is a method for evaluating the quenched average of the Bethe free energy over the random fields:
\begin{align}
[F_{\mrm{bethe}}(\bm{h},\beta)]_{\bm{h}} := \int \diff \bm{h} \Big(\prod_{i \in V}p_i(h_i)\Big) F_{\mrm{bethe}}(\bm{h},\beta),
\label{eq:quenched-BetheFreeEnergy}
\end{align}
where the notation $[\cdots]_{\bm{h}}$ represents the average value over the random fields, and 
$p_i(h_i)$ is the distribution of the field on vertex $i$, where these distributions can vary by vertex in general.

\section{Proposed Method} \label{sec:ProposeMethod}

\begin{figure}[htb]
\begin{center}
\includegraphics[height=4.0cm]{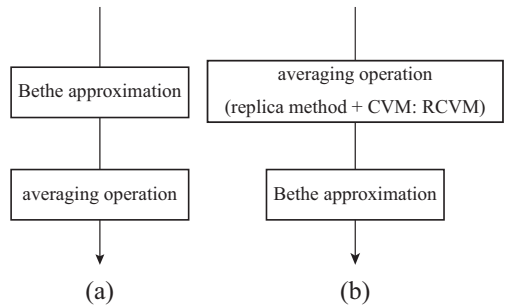}
\end{center}
\caption{Procedures of obtaining the quenched average of the Bethe free energy. (a) Principled procedure. (b) Procedure of the proposed method based on the RCVM.}
\label{fig:SchemeRLBP}
\end{figure}

According to equation (\ref{eq:quenched-BetheFreeEnergy}), in principle, 
we have to perform the averaging operation after constructing the Bethe free energy to obtain $[F_{\mrm{bethe}}]_{\bm{h}}$ (see figure \ref{fig:SchemeRLBP}(a)).  
However, it is not straightforward to directly integrate the Bethe free energy. 
Thus, we adopt another strategy.

In this paper, we propose an approximate method based on the idea of the RCVM~\cite{RCVM2010, RCVM2013}.
Figure \ref{fig:SchemeRLBP}(b) shows the procedure of our method. 
In the method, we first take the average of the free energy using the replica method and CVM, namely, the RCVM (section \ref{sec:RLBP}), 
and then, we apply the Bethe approximation to the resulting form of the RCVM (section \ref{sec:BetheApprox}). 
If the exchange of the order of the two operations, the Bethe approximation and the quenched averaging operation, is allowed, 
we can expect to obtain a good approximation of the quenched average of the Bethe free energy.
  
\subsection{Replica Cluster Variation Method} \label{sec:RLBP}

\begin{figure*}[htb]
\begin{center}
\includegraphics[height=5.0cm]{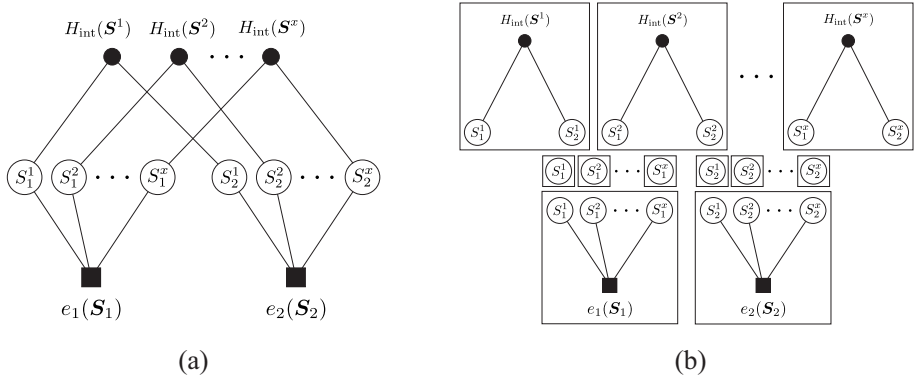}
\end{center}
\caption{(a) Factor graph representation of $P_x(\mcal{S}_x)$ when $n = 2$. (b) Cluster decomposition for (a). In the decomposition, the three different types of clusters are employed: the clusters consist of $S_i^{\alpha}$, of $\bm{S}_i$, and of $\bm{S}^{\alpha}$.}
\label{fig:CVM-decomp}
\end{figure*}

First, we obtain the quenched average of the true free energy of the MRF in equation (\ref{eq:MRF}), that is, 
\begin{align*}
[F(\bm{h},\beta)]_{\bm{h}}=-\frac{1}{\beta}\int \diff \bm{h}\Big( \prod_{i \in V}p_i(h_i)\Big) \ln Z(\bm{h},\beta).
\end{align*}
In the context of the replica method~\cite{ParisiBook1987,Nishimori2001}, we have
\begin{align}
[F(\bm{h},\beta)]_{\bm{h}}=-\frac{1}{\beta}\lim_{x\to 0}\frac{Z_x - 1}{x},
\label{eq:replica-method}
\end{align}
where
\begin{align*}
Z_x := \int \diff \bm{h} \Big(\prod_{i \in V}p_i(h_i)\Big) Z(\bm{h},\beta)^x.
\end{align*}
By assuming that $x$ is a natural number, we obtain
\begin{align*}
Z_x = \sum_{\mcal{S}_x}\exp \beta \Big(\sum_{ i\in V}e_i(\bm{S}_i)
+ \sum_{\{i,j\} \in E} \sum_{\alpha=1}^x \psi_{i.j}(S_i^{\alpha}, S_j^{\alpha})\Big),
\end{align*}
where $\mcal{S}_x = \{S_i^{\alpha} \mid i \in V , \> \alpha = 1,2,\ldots, x\}$,  
$\bm{S}_i =\{ S_i^{\alpha} \mid \alpha = 1,2,\ldots, x\}$, and the function $e_i(\bm{S}_i)$ is defined by
\begin{align*}
e_i(\bm{S}_i):=\frac{1}{\beta}\ln \int \diff h\, p_i(h) \exp\Big(\beta \sum_{\alpha=1}^x \phi_i(S_i^{\alpha}, h)\Big).
\end{align*}
We regard $\mcal{Z}_x$ as the partition of the $x$-replicated system and define the $x$-replicated free energy as
\begin{align}
F_x &:= -\frac{1}{\beta} \ln Z_x\nn
&\>= -\sum_{i \in V}\sum_{\bm{S}_i}e_i(\bm{S}_i) Q_i(\bm{S}_i) + \sum_{\alpha = 1}^x\sum_{\bm{S}^{\alpha}}H_{\mrm{int}}(\bm{S}^{\alpha})Q^{\alpha}(\bm{S}^{\alpha})\nn
\aldef
+\frac{1}{\beta}\sum_{\mcal{S}_x}P_x(\mcal{S}_x)\ln P_x(\mcal{S}_x),
\label{eq:x-replicatedFreeEnergy}
\end{align}
where
\begin{align*}
P_x(\mcal{S}_x)&:=\frac{1}{Z_x}\exp \beta \Big(\sum_{ i\in V}e_i(\bm{S}_i)
-\sum_{\alpha = 1}^xH_{\mrm{int}}(\bm{S}^{\alpha})\Big)
\end{align*}
is the Gibbs distribution of the $x$-replicated system, and $\bm{S}^{\alpha} = \{S_i^{\alpha} \mid i \in V\}$. 
The energy function $H_{\mrm{int}}(\bm{S})$ is defined by 
\begin{align*}
H_{\mrm{int}}(\bm{S}):= - \sum_{\{i,j\} \in E}\psi_{i.j}(S_i, S_j)
\end{align*}
and is just the interaction term of the original system. 
The distributions, $Q_i(\bm{S}_i)$ and $Q^{\alpha}(\bm{S}^{\alpha})$, are the marginal distributions of the distribution $P_x(\mcal{S}_x)$.
The factor graph representation of the $x$-replicated system is shown in figure \ref{fig:CVM-decomp}(a).

In accordance with the cluster decomposition based on the CVM shown in figure \ref{fig:CVM-decomp}(b),
by using the marginal distributions, $\{Q_i(\bm{S}_i),Q^{\alpha}(\bm{S}^{\alpha})\}$, together with the one-variable marginal distributions of $P_x(\mcal{S}_x)$, $\{Q_i^{\alpha}(S_i^{\alpha})\}$, 
we approximate the Gibbs distribution of the $x$-replicated system as
\begin{widetext}
\begin{align}
P_x(\mcal{S}_x)\approx \frac{\big(\prod_{i \in V}\prod_{\alpha=1}^x Q_{i}^{\alpha} (S_i^{\alpha}) \big)\big(\prod_{i \in V}Q_i(\bm{S}_i)\big)\big( \prod_{\alpha=1}^x Q^{\alpha}(\bm{S}^{\alpha})\big)}{\big(\prod_{i \in V}\prod_{\alpha=1}^x Q_{i}^{\alpha} (S_i^{\alpha}) \big)^2}
=\frac{\big(\prod_{i \in V}Q_i(\bm{S}_i)\big)\big( \prod_{\alpha=1}^x Q^{\alpha}(\bm{S}^{\alpha})\big)}{\prod_{i \in V}\prod_{\alpha=1}^x Q_{i}^{\alpha} (S_i^{\alpha}) }.
\label{eq:CVM}
\end{align}
\end{widetext}
As in equation (\ref{eq:BetheDecomp-original}), $P_x(\mcal{S}_x)$ is approximated by the product of the marginal distributions of the clusters 
divided by the product of the marginal distributions of the double-counted clusters.
By applying this approximation to $P_x(\mcal{S}_x)$ in the logarithmic function in the last term in equation (\ref{eq:x-replicatedFreeEnergy}), 
we obtain the expression of the variational free energy as
\begin{align}
\mcal{F}_x^{\mrm{RCVM}}[\{ Q_i, Q^{\alpha}, Q_i^{\alpha}\}] &:= \sum_{i \in V}\mcal{V}_i[Q_i] + \sum_{\alpha = 1}^x \mcal{V}_{\mrm{int}}[Q^{\alpha}]\nn
\aldef
-\frac{1}{\beta}\sum_{i \in V}\sum_{\alpha=1}^x \mcal{H}_1[Q_{i}^{\alpha}],
\label{eq:VariationalFreeEnergy}
\end{align}
where the functionals, $\mcal{V}_i[Q_i]$ and $\mcal{V}_{\mrm{int}}[Q^{\alpha}]$, are defined as
\begin{align}
\mcal{V}_i[Q_i]&:=-\sum_{\bm{S}_i}e_i(\bm{S}_i)Q_i(\bm{S}_i) +\frac{1}{\beta}\sum_{\bm{S}_i}Q_i(\bm{S}_i)\ln Q_i(\bm{S}_i),
\label{eq:def-Vi}\\
\mcal{V}_{\mrm{int}}[Q^{\alpha}]&:=\sum_{\bm{S}^{\alpha}}H_{\mrm{int}}(\bm{S}^{\alpha})
Q^{\alpha}(\bm{S})
+\frac{1}{\beta}\sum_{\bm{S}^{\alpha}}Q^{\alpha}(\bm{S}^{\alpha}) \ln Q^{\alpha}(\bm{S}^{\alpha}).
\label{eq:def-V^alpha}
\end{align}
In the context of the CVM, the $x$-replicated free energy in equation (\ref{eq:x-replicatedFreeEnergy}) is approximated by 
the minimum of the variational free energy in equation (\ref{eq:VariationalFreeEnergy}) with respect to the marginal distributions $\{ Q_i, Q^{\alpha}, Q_i^{\alpha}\}$, i.e., 
$F_x \approx \min_{\{ Q_i, Q^{\alpha}, Q_i^{\alpha}\}} \mcal{F}_x^{\mrm{RCVM}}[\{ Q_i, Q^{\alpha}, Q_i^{\alpha}\}]$.
Note that, at the minimum point, the normalization constraints for $\{ Q_i, Q^{\alpha}, Q_i^{\alpha}\}$ and the marginal constraints,
\begin{align}
Q_{i}^{\alpha} (S_i^{\alpha}) &= \sum_{\bm{S}_i \setminus \{S_{i}^{\alpha}\}}Q_i(\bm{S}_i)
\label{eq:marginal-constraints_Qi}
\end{align}
and
\begin{align}
Q_{i}^{\alpha} (S_i^{\alpha}) &= \sum_{\bm{S}^{\alpha} \setminus \{S_{i}^{\alpha}\}}Q^{\alpha}(\bm{S}^{\alpha}),
\label{eq:marginal-constraints_Q^alpha}
\end{align}
should hold.

In order to minimize the variational free energy with respect to $\{Q_i(\bm{S}_i)\}$, by using the Lagrange multipliers, 
we perform the variational minimization of
\begin{align*}
&\mcal{L}[\{Q_i\}]:= \sum_{i \in V}\mcal{V}_i[Q_i] - \sum_{i \in V}a_i \Big(\sum_{\bm{S}_i}Q_i(\bm{S}_i) - 1\Big)\nn
&-\sum_{i \in V}\sum_{\alpha = 1}^x\sum_{S_i^{\alpha}}\Lambda_i^{\alpha}(S_i^{\alpha})\Big( \sum_{\bm{S}_i \setminus \{S_{i}^{\alpha}\}}Q_i(\bm{S}_i) - Q_{i}^{\alpha} (S_i^{\alpha})\Big)
\end{align*}
with respect to $\{Q_i(\bm{S}_i)\}$. From the result of this minimization, we obtain
\begin{align}
Q_i(\bm{S}_i)\propto \exp \beta \Big(e_i(\bm{S}_i) + \sum_{\alpha = 1}^x\Lambda_i^{\alpha}(S_i^{\alpha})\Big).
\label{eq:Qi}
\end{align}
The Lagrange multipliers, $\bm{\Lambda}_i = \{\Lambda_i^{\alpha}(S_i) \mid \alpha = 1,2,\ldots, x\}$, are determined such that they satisfy equation (\ref{eq:marginal-constraints_Qi}). 
By substituting equation (\ref{eq:Qi}) into equation (\ref{eq:def-Vi}), while noting the marginal constraints in equation (\ref{eq:marginal-constraints_Qi}), 
we obtain the partially minimized variational free energy, 
\begin{align*}
\mcal{F}_x^{\mrm{RCVM}}[\{Q^{\alpha}, Q_i^{\alpha}\}]:=\min_{\{Q_i\}}\mcal{F}_x^{\mrm{RCVM}}[\{ Q_i, Q^{\alpha}, Q_i^{\alpha}\}],
\end{align*}
as
\begin{align}
&\mcal{F}_x^{\mrm{RCVM}}[\{Q^{\alpha}, Q_i^{\alpha}\}]
 = \sum_{i \in V}\extr{\bm{\Lambda}_i}\Big\{\sum_{\alpha = 1}^x\sum_{S_i^{\alpha}}\Lambda_i^{\alpha}(S_i^{\alpha})Q_i^{\alpha}(S_i^{\alpha})\nn
&-\frac{1}{\beta}\ln \int \diff h\, p_i(h)\prod_{\alpha = 1}^x\sum_{S_i^{\alpha}} \exp \beta \big( \phi_i(S_i^{\alpha},h) \nn
&+ \Lambda_i^{\alpha}(S_i^{\alpha})\big)\Big\} + \sum_{\alpha = 1}^x \mcal{V}_{\mrm{int}}[Q^{\alpha}]
-\frac{1}{\beta}\sum_{i \in V}\sum_{\alpha=1}^x \mcal{H}_1[Q_{i}^{\alpha}],
\label{eq:VariationalFreeEnergy-reduce}
\end{align}
where the notation ``extr'' denotes the extremum with respect to the assigned parameters. 

\subsection{Bethe Approximation and Replica Symmetric Ansatz} \label{sec:BetheApprox}

The functional $\mcal{V}_{\mrm{int}}[Q^{\alpha}]$ can be interpreted as the variational free energy for the interaction term of the original system (see equation (\ref{eq:def-V^alpha})). 
Since this variational free energy is intractable in general, we approximate it by the Bethe approximation. 
As in equation (\ref{eq:BetheDecomp-original}), we approximate $Q^{\alpha}(\bm{S}^{\alpha})$ by
\begin{align}
Q^{\alpha}(\bm{S}^{\alpha})\approx  
\frac{\big(\prod_{i \in V}Q_i^{\alpha}(S_i^{\alpha})\big)\big(\prod_{\{i,j\} \in E} Q_{i,j}^{\alpha}(S_i^{\alpha}, S_j^{\alpha})\big)  }{\prod_{\{i,j\} \in E}Q_i^{\alpha}(S_i^{\alpha})Q_j^{\alpha}(S_j^{\alpha})},
\label{eq:BetheApproximation-V^alpha}
\end{align}
where the distributions, $Q_i^{\alpha}(S_i^{\alpha})$ and $Q_{i,j}^{\alpha}(S_i^{\alpha}, S_j^{\alpha})$, 
are the marginal distributions of $Q^{\alpha}(\bm{S}^{\alpha})$, so that the marginal constraints, 
\begin{align*}
\sum_{S_i^{\alpha}}Q_{i,j}^{\alpha}(S_i^{\alpha}, S_j^{\alpha})&=Q_j^{\alpha}(S_j^{\alpha})
\end{align*}
and
\begin{align*}
\sum_{S_j^{\alpha}}Q_{i,j}^{\alpha}(S_i^{\alpha}, S_j^{\alpha})&=Q_i^{\alpha}(S_i^{\alpha}),
\end{align*}
are satisfied. By applying equation (\ref{eq:BetheApproximation-V^alpha}) to 
$Q^{\alpha}(\bm{S}^{\alpha})$ in the logarithmic function in the last term of equation (\ref{eq:def-V^alpha}), we obtain the Bethe approximation of $\mcal{V}_{\mrm{int}}[Q^{\alpha}]$ as
\begin{align}
&
\mcal{V}_{\mrm{int}}^{\mrm{bethe}}[\{Q_i^{\alpha},Q_{i,j}^{\alpha}\}]\nn
&:=-\sum_{\{i,j\} \in E}\sum_{S_i^{\alpha},S_j^{\alpha}}\psi_{i,j}(S_i^{\alpha},S_j^{\alpha})Q_{i,j}^{\alpha}(S_i^{\alpha},S_j^{\alpha})
\nn
&+\frac{1}{\beta}\sum_{i \in V}\mcal{H}_1[Q_i^{\alpha}]+\frac{1}{\beta}\sum_{\{i,j\} \in E}\big(\mcal{H}_2[Q_{i,j}^{\alpha}] - \mcal{H}_1[Q_i^{\alpha}] - \mcal{H}_1[Q_j^{\alpha}]\big).
\label{eq:BetheFreeEnergy-V^alpha}
\end{align}

By substituting the Bethe approximation in equation (\ref{eq:BetheFreeEnergy-V^alpha}) into equation (\ref{eq:VariationalFreeEnergy-reduce}), we obtain the Bethe approximation 
of equation (\ref{eq:VariationalFreeEnergy-reduce}): $\mcal{F}_x^{\mrm{RCVM}}[\{Q^{\alpha}, Q_i^{\alpha}\}]\approx \mcal{F}_x^{\mrm{LBP}}[\{Q_i^{\alpha}, Q_{i,j}^{\alpha}\}]$.
After the Bethe approximation, we make the replica symmetric (RS) assumption~\cite{ParisiBook1987,Nishimori2001} in $\mcal{F}_x^{\mrm{LBP}}[\{Q_i^{\alpha}, Q_{i,j}^{\alpha}\}]$, and subsequently, by taking the limit as $x \to 0$, we finally reach the variational free energy expressed as
\begin{align}
&\mcal{F}^{\mrm{LBP(RS)}}[\{Q_i, Q_{i,j}\}]
= \sum_{i \in V}\extr{\Lambda_i}\Big\{ \sum_{S_i}\Lambda_i(S_i)Q_i(S_i)\nn
&-\frac{1}{\beta}\int \diff h\, p_i(h)\ln \sum_{S_i} \exp \beta \big( \phi_i(S_i,h) 
+ \Lambda_i(S_i)\big)\Big\}\nn
& -\sum_{\{i,j\} \in E}\sum_{S_i,S_j}\psi_{i,j}(S_i,S_j)Q_{i,j}(S_i,S_j)\nn
&+\frac{1}{\beta}\sum_{\{i,j\} \in E}\big(\mcal{H}_2[Q_{i,j}] - \mcal{H}_1[Q_i] - \mcal{H}_1[Q_j]\big).
\label{eq:quenched-FreeEnergy-Bethe(RS)}
\end{align}
The detailed derivation of this variational free energy is shown in appendix \ref{app:sec:Deriv-BetheFreeEnergy-RS}.
We expect that the minimum of this variational free energy is the approximation of the quenched average of the Bethe free energy in equation (\ref{eq:quenched-BetheFreeEnergy}): 
$[F_{\mrm{bethe}}(\bm{h},\beta)]_{\bm{h}} \approx F_{\mrm{min}}^{\mrm{LBP(RS)}}$,
where
\begin{align}
F_{\mrm{min}}^{\mrm{LBP(RS)}}:=\min_{\{Q_i, Q_{i,j}\}}\mcal{F}^{\mrm{LBP(RS)}}[\{Q_i, Q_{i,j}\}].
\label{eq:FreeEnergy-Bethe(RS)_min}
\end{align}
At the minimum of the variational free energy, $\mcal{F}^{\mrm{LBP(RS)}}[\{Q_i, Q_{i,j}\}]$, the normalization constraints, 
\begin{align}
\sum_{S_i}Q_i(S_i) = \sum_{S_i,S_j}Q_{i,j}(S_i,S_j) = 1,
\label{eq:normalization-constraints-Bethe(RS)}
\end{align}
and the marginal constraints, 
\begin{align}
\sum_{S_i}Q_{i,j}(S_i, S_j)=Q_j(S_j) 
\label{eq:marginal-constraints-Bethe(RS)-i}
\end{align}
and
\begin{align}
\sum_{S_j}Q_{i,j}(S_i, S_j)=Q_i(S_i), 
\label{eq:marginal-constraints-Bethe(RS)-j}
\end{align}
should hold.

\subsection{Message-passing Equation}
\label{sec:ReplicaMessagePassing}

In this section, we show the message-passing equation for minimizing the variational free energy in equation (\ref{eq:quenched-FreeEnergy-Bethe(RS)}) obtained in the previous section.

The message-passing equation for our method is obtained as
\begin{align}
\mu_{j \to i}(S_i) 
&\propto \sum_{S_j}Q_j(S_j)\exp \big( \beta\psi_{i,j}(S_i,S_j)\big)\mcal\mu_{i \to j}(S_j)^{-1},
\label{eq:MessagePassing-LBP(RS)}\\
Q_i(S_i) &= \int \diff h\, p_i(h) \frac{\exp \beta \big( \phi_i(S_i,h) + \Lambda_i(S_i)\big)}{ \sum_{S_i} \exp \beta \big( \phi_i(S_i,h) + \Lambda_i(S_i)\big)},
\label{eq:Qi-LBP(RS)}\\
\beta \Lambda_i(S_i) &= \sum_{k \in \partial i } \ln \mu_{k \to i}(S_i).
\label{eq:exp(Lambda)}
\end{align}
The quantity $\mu_{i \to j}(S_j)$ is the message from vertex $i$ to vertex $j$, 
and the Lagrange multipliers $\{\Lambda_i(S_i)\}$ in equation (\ref{eq:exp(Lambda)}) satisfy the extremal conditions 
in the first term in equation (\ref{eq:quenched-FreeEnergy-Bethe(RS)}).
By using the messages and $\{Q_i(S_i)\}$, the two-vertex marginal distributions $\{Q_{i,j}(S_i,S_j)\}$ are obtained as
\begin{align}
Q_{i,j}(S_i,S_j) &\propto Q_i(S_i)Q_j(S_j)\exp \big(\beta \psi_{i,j}(S_i,S_j)\big)\nn
\aleq \times \mu_{j \to i}(S_i)^{-1}\mu_{i \to j}(S_j)^{-1}.
\label{eq:Qij-LBP(RS)}
\end{align}
The detailed derivation of equations (\ref{eq:MessagePassing-LBP(RS)})--(\ref{eq:Qij-LBP(RS)}) is shown in appendix \ref{app:sec:message-passing}.  
The order of the computational cost of  the proposed message-passing equation is equivalent to 
that of standard LBP in section \ref{sec:LBP-detail}.

When the distributions of the random fields are Dirac delta functions, $p_i(h) = \delta(h - h_i)$, 
namely, the fields are not the quenched parameters but the fixed parameters,
the method presented in equations (\ref{eq:MessagePassing-LBP(RS)})--(\ref{eq:Qij-LBP(RS)}) is reduced to the standard LBP in equations (\ref{eq:message-passing})--(\ref{eq:LBP-bij}).
In this case, $F_{\mrm{min}}^{\mrm{LBP(RS)}}$ in equation (\ref{eq:FreeEnergy-Bethe(RS)_min}) is equivalent to the Bethe free energy $F_{\mrm{bethe}}(\bm{h}, \beta)$.

After numerically solving the simultaneous equations in equations (\ref{eq:MessagePassing-LBP(RS)})--(\ref{eq:Qij-LBP(RS)}), 
by substituting the solutions, $\{Q_i(S_i)\}$, $\{Q_{i,j}(S_i,S_j)\}$, and $\{\Lambda_i(S_i)\}$, into equation (\ref{eq:quenched-FreeEnergy-Bethe(RS)}), 
we obtain the minimum values of the variational free energy, $F_{\mrm{min}}^{\mrm{LBP(RS)}}$, 
and regard it as the approximation of $[F_{\mrm{bethe}}(\bm{h}, \beta)]_{\bm{h}}$ in equation (\ref{eq:quenched-BetheFreeEnergy}).

For the moment, we suppose that the function $\phi_i(S_i, h_i)$ can be divided as $\phi_i(S_i, h_i) = \phi_i^{(0)}(S_i, h_i) + \phi_i^{(1)}(S_i)$. 
The variations in the quenched average of the Bethe free energy in equation (\ref{eq:quenched-BetheFreeEnergy}) with respect to $\phi_i^{(1)}(S_i)$ and $\psi_{i,j}(S_i,S_j)$ are 
\begin{align}
\frac{\delta [F_{\mrm{bethe}}]_{\bm{h}} }{\delta \phi_i^{(1)}(S_i)}= -[b_i(S_i)]_{\bm{h}}
\label{eq:variations-[F_bethe]_h-1}
\end{align}
and
\begin{align}
\frac{\delta [F_{\mrm{bethe}}]_{\bm{h}}}{\delta \psi_{i,j}(S_i,S_j)} = -[b_{i,j}(S_i,S_j)]_{\bm{h}},
\label{eq:variations-[F_bethe]_h-2}
\end{align}
respectively, which are the quenched average of the beliefs obtained from LBP.
On the other hand, the variations in $F_{\mrm{min}}^{\mrm{LBP(RS)}}$  
with respect to $\phi_i^{(1)}(S_i)$ and $\psi_{i,j}(S_i,S_j)$ are obtained as
\begin{align}
\frac{\delta F_{\mrm{min}}^{\mrm{LBP(RS)}}}{\delta \phi_i^{(1)}(S_i)}=-Q_i(S_i)
\label{eq:variations-F_LBP(RS)-1}
\end{align}
and
\begin{align}
\frac{\delta F_{\mrm{min}}^{\mrm{LBP(RS)}}}{\delta \psi_{i,j}(S_i,S_j)}=-Q_{i,j}(S_i,S_j),
\label{eq:variations-F_LBP(RS)-2}
\end{align}
respectively. By comparing equations (\ref{eq:variations-[F_bethe]_h-1}) and (\ref{eq:variations-[F_bethe]_h-2}) with equations (\ref{eq:variations-F_LBP(RS)-1}) and (\ref{eq:variations-F_LBP(RS)-2}), 
it can be expected that, if $F_{\mrm{min}}^{\mrm{LBP(RS)}}$ is a good approximation of $[F_{\mrm{bethe}}(\bm{h}, \beta)]_{\bm{h}}$, 
the marginal distributions, $Q_i(S_i)$ and $Q_{i,j}(S_i,S_j)$, are also good approximations of 
the quenched averages of the beliefs, $[b_i(S_i)]_{\bm{h}}$ and $[b_{i,j}(S_i,S_j)]_{\bm{h}}$, respectively.

\subsection{Numerical Experiment} \label{sec:check-validity}

In this section, we describe the evaluation of the validity of our method by using numerical experiments. 
In the experiments, we used the model expressed as 
\begin{align}
P(\bm{S}\mid \bm{h}) = \frac{1}{Z(\bm{h})}\exp  \Big( \sum_{i \in V} h_i S_i + \sum_{\{i,j\} \in E}J_{ij}S_i S_j\Big),
\label{eq:EA-Model}
\end{align}
which is defined on a certain graph, where $S_i$ takes $q$ real values in the interval $[-1,1]$ as $S_i \in \{2 S /(q-1) - 1 \mid S = 0,1,2,\ldots, q-1\}$, 
i.e., $S_i \in \{ +1,-1\}$ when $q = 2$, $S_i \in \{ +1,0,-1\}$ when $q = 3$, and so on.
The fields $\bm{h}$ are i.i.d. random fields drawn from the Gaussian distribution with a mean of zero and a variance of $\sigma^2$, 
$\mcal{N}(h_i \mid 0, \sigma^2)$.

We compared the free energy per variable obtained by our method, $F_{\mrm{min}}^{\mrm{LBP(RS)}} / n$, 
with the quenched average of the Bethe free energy (per variable) shown in equation (\ref{eq:quenched-BetheFreeEnergy}), 
which was obtained by numerically averaging the Bethe free energy in equation (\ref{eq:BetheFreeEnergy}) over the random fields, 
and we compared the behaviors of the quenched average of the magnetizations obtained by the two different methods. 
The magnetizations obtained from LBP and our method are given by
$M_{\mrm{LBP}}:=n^{-1}\sum_{i \in V}\sum_{S_i}S_i [b_i(S_i)]_{\bm{h}}$ and 
$M_{\mrm{RLBP}}:=n^{-1}\sum_{i \in V}\sum_{S_i}S_i Q_i(S_i)$,
respectively.

\subsubsection{Square Lattice}
\label{sec:munerical-square}

\begin{figure*}[htb]
\begin{center}
\includegraphics[height=4.5cm]{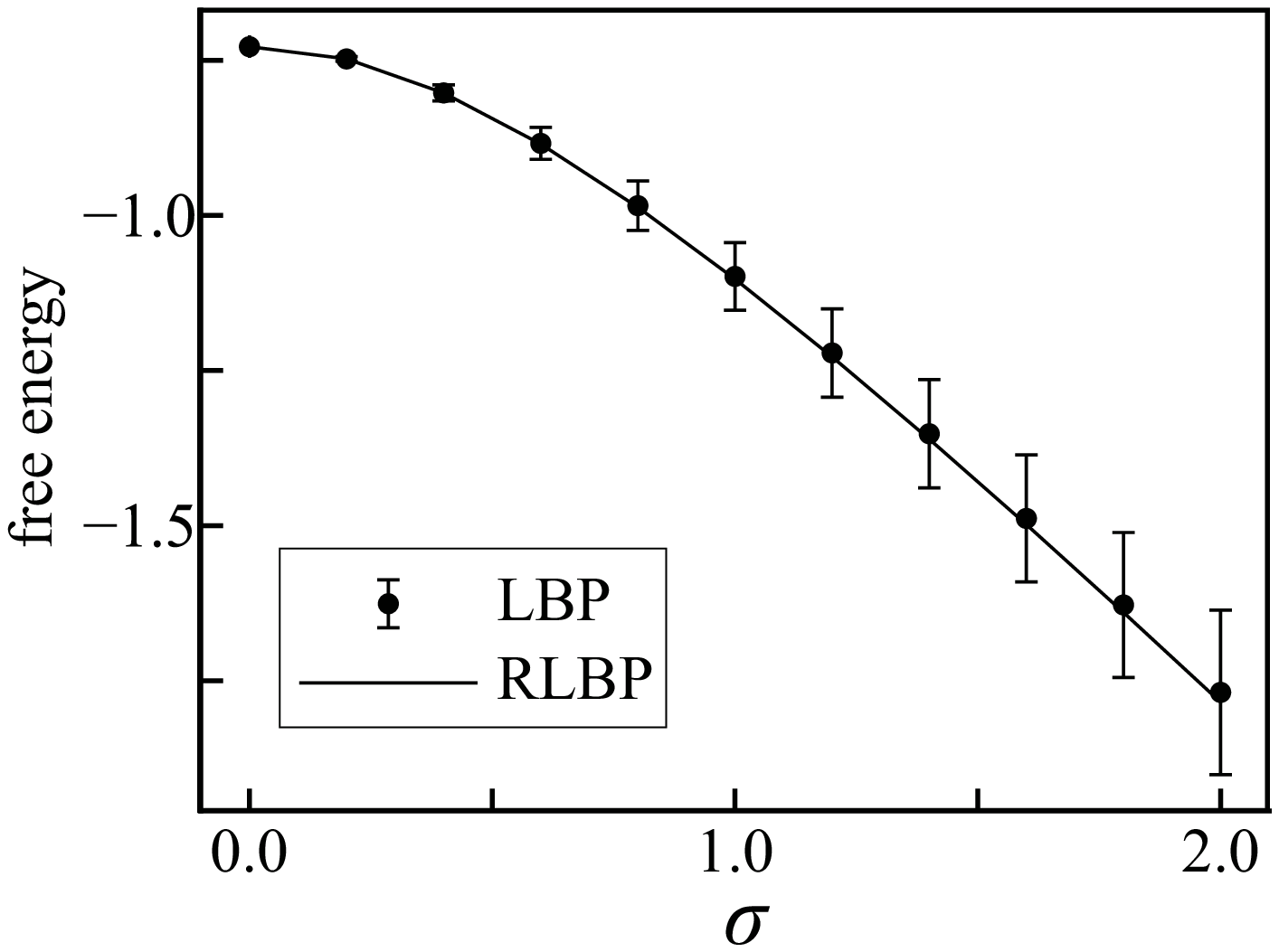}
\includegraphics[height=4.5cm]{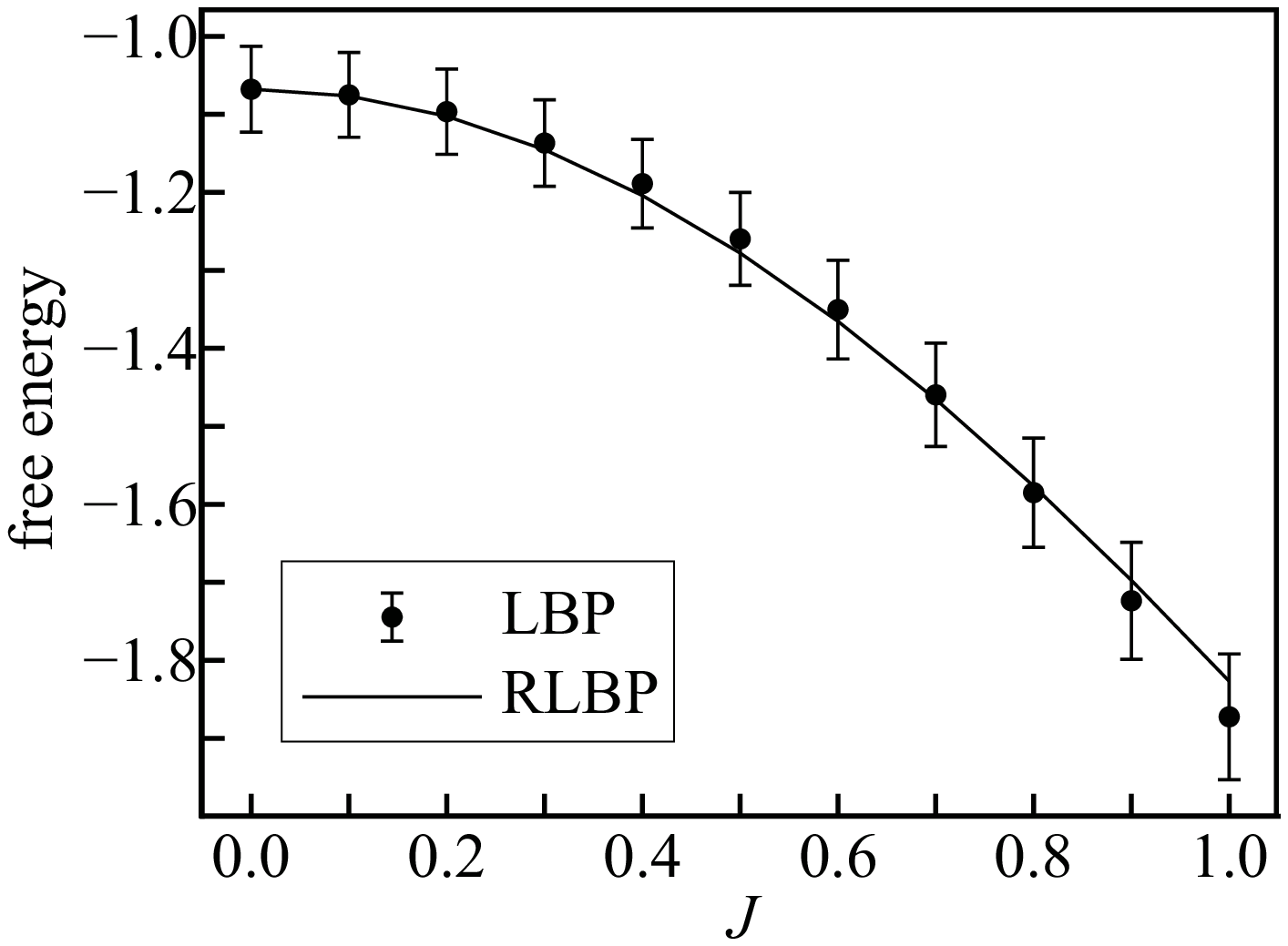}
\end{center}
\caption{Quenched Bethe free energies per variable for $q = 2$. The left panel shows the free energies versus $\sigma$ with $J = 0.2$, and the right panel shows the free energies versus $J$ with $\sigma = 1$. The error bars are the standard deviation.}
\label{fig:8grid-Q2}
\end{figure*}

\begin{figure*}[htb]
\begin{center}
\includegraphics[height=4.5cm]{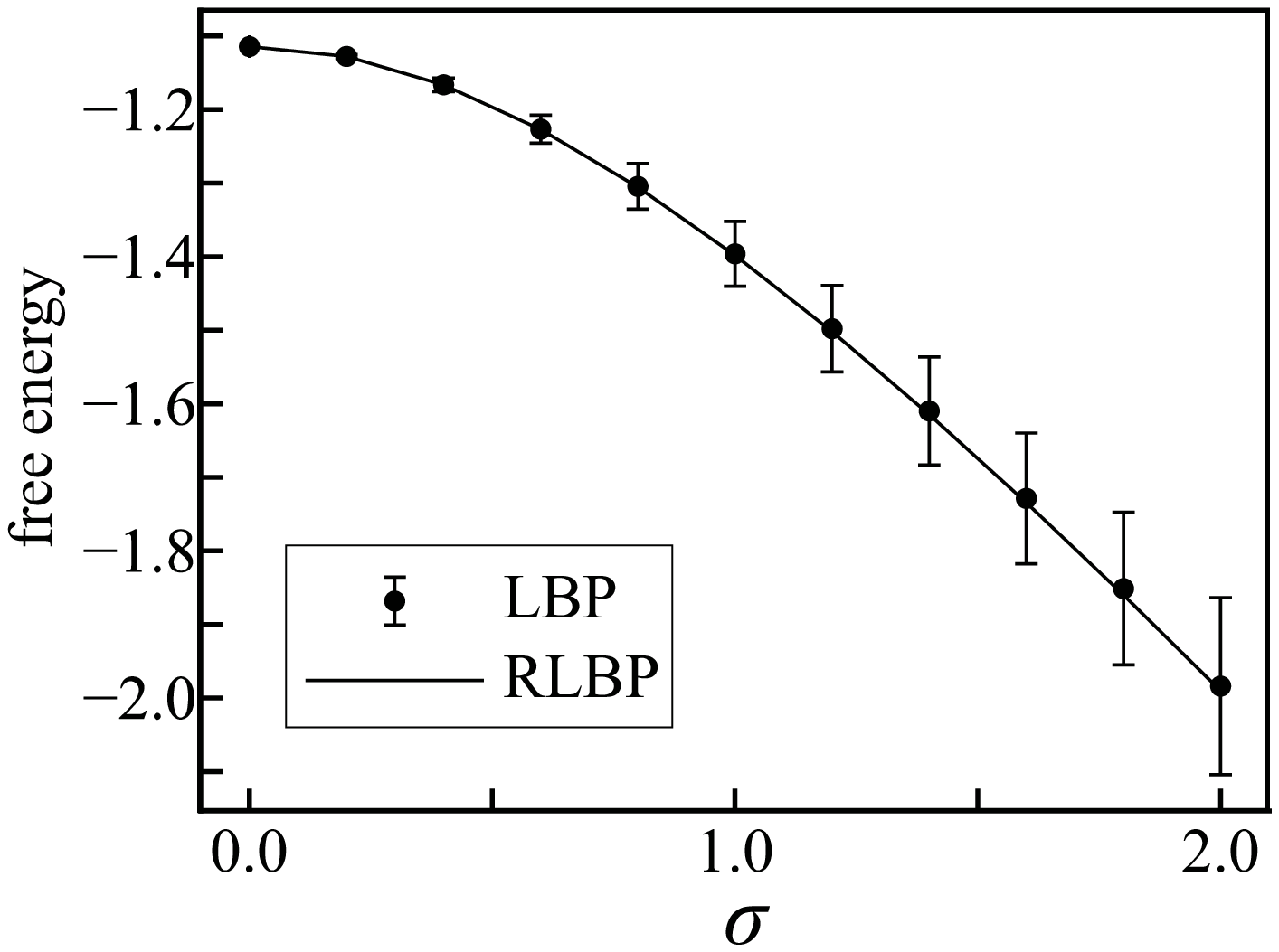}
\includegraphics[height=4.5cm]{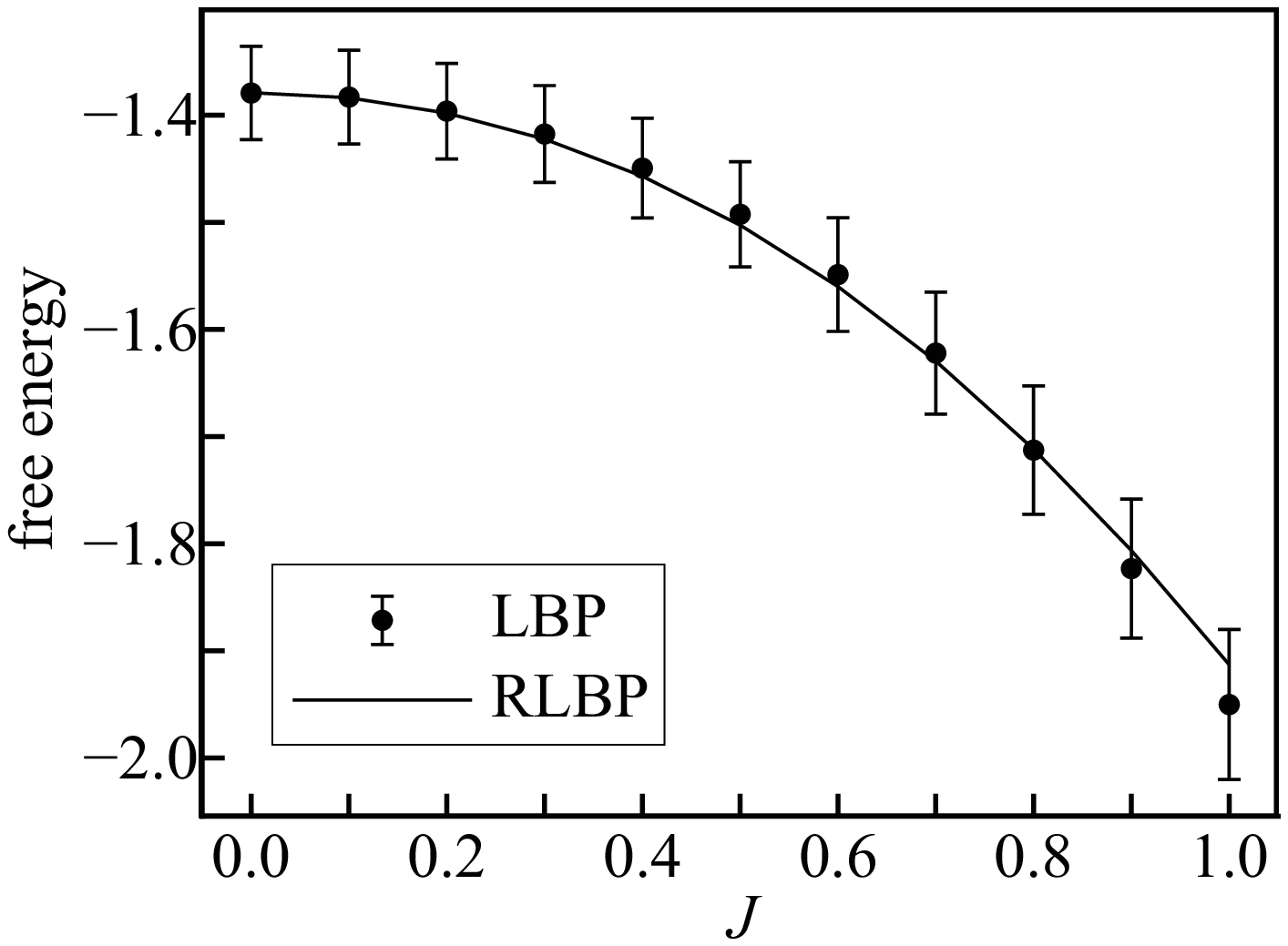}
\end{center}
\caption{Quenched Bethe free energies per variable for $q = 3$. The left panel shows the free energies versus $\sigma$ with $J = 0.2$, and the right panel shows the free energies versus $J$ with $\sigma = 1$. The error bars are the standard deviation.}
\label{fig:8grid-Q3}
\end{figure*}

\begin{figure*}[htb]
\begin{center}
\includegraphics[height=4.5cm]{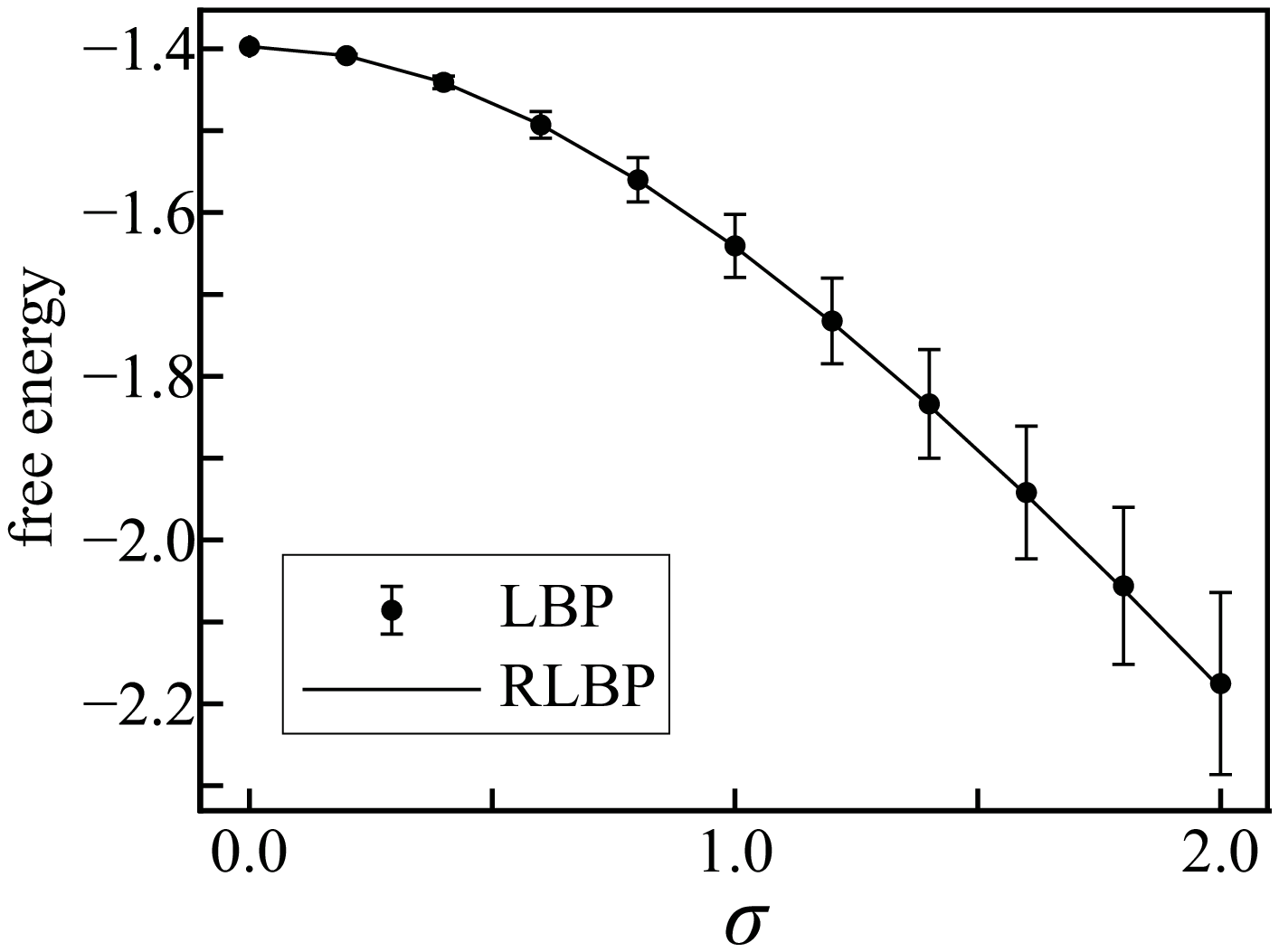}
\includegraphics[height=4.5cm]{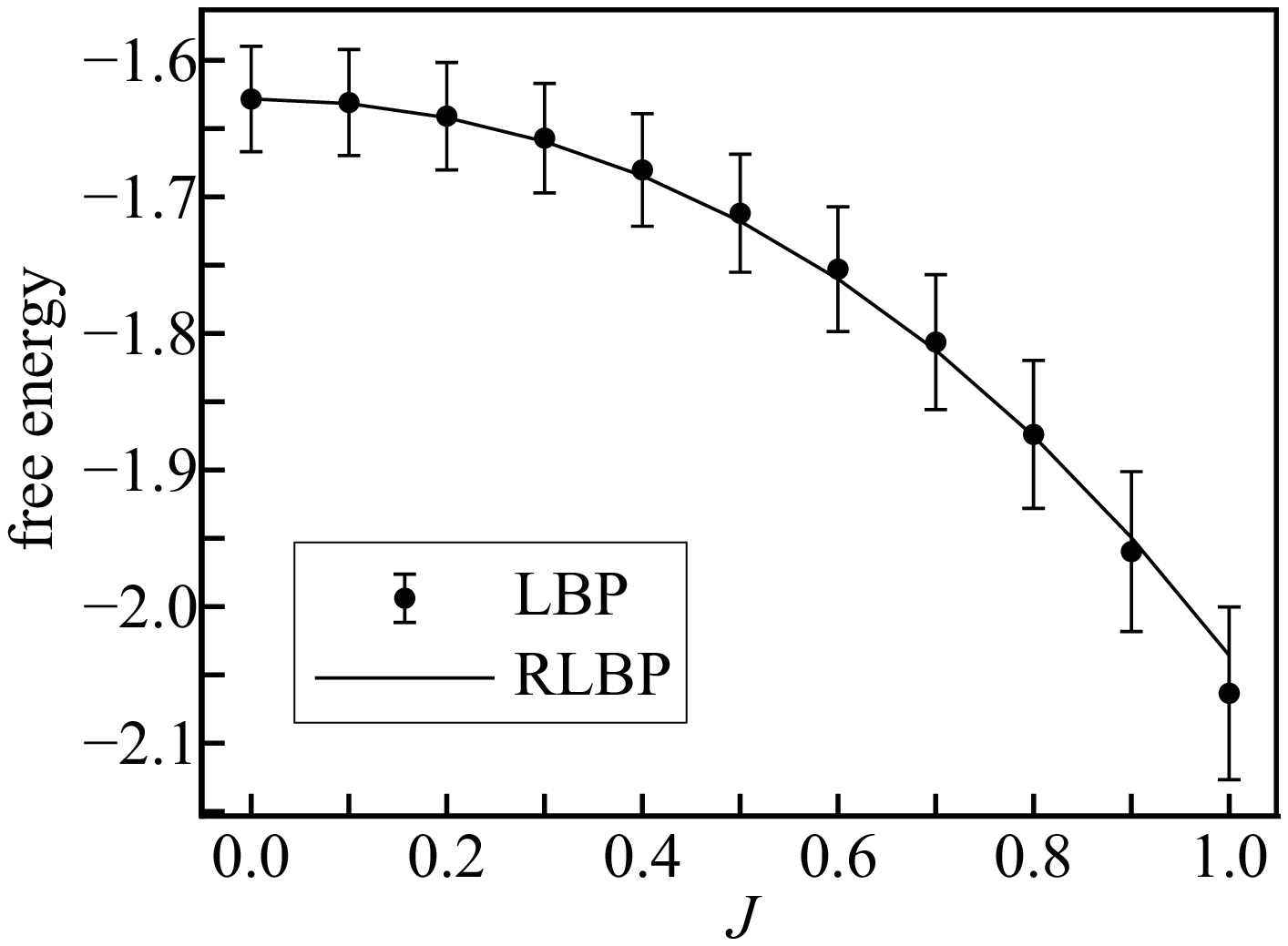}
\end{center}
\caption{Quenched Bethe free energies per variable for $q = 4$. The left panel shows the free energies versus $\sigma$ with $J = 0.2$, and the right panel shows the free energies versus $J$ with $\sigma = 1$. The error bars are the standard deviation.}
\label{fig:8grid-Q4}
\end{figure*}

We show the results when the model in equation (\ref{eq:EA-Model}) is defined on a graph of an $8 \times 8$ square lattice with the free boundary condition and when all of the interactions are unique, $J_{ij} = J$.
Figures \ref{fig:8grid-Q2}--\ref{fig:8grid-Q4} show the plots for $q = 2$, $q = 3$, and $q = 4$. 
``LBP'' represents the results obtained by the numerically averaged Bethe free energy, and ``RLBP'' represents the results obtained by our method.
Each plot of LBP is numerically averaged over 10000 realizations of the random fields.
In almost all cases, the results of our method are consistent with the numerically averaged Bethe free energies, as expected. 
However, in the cases of large $J$, mismatches between the two methods are observed.

\begin{figure}[htb]
\begin{center}
\includegraphics[height=4.5cm]{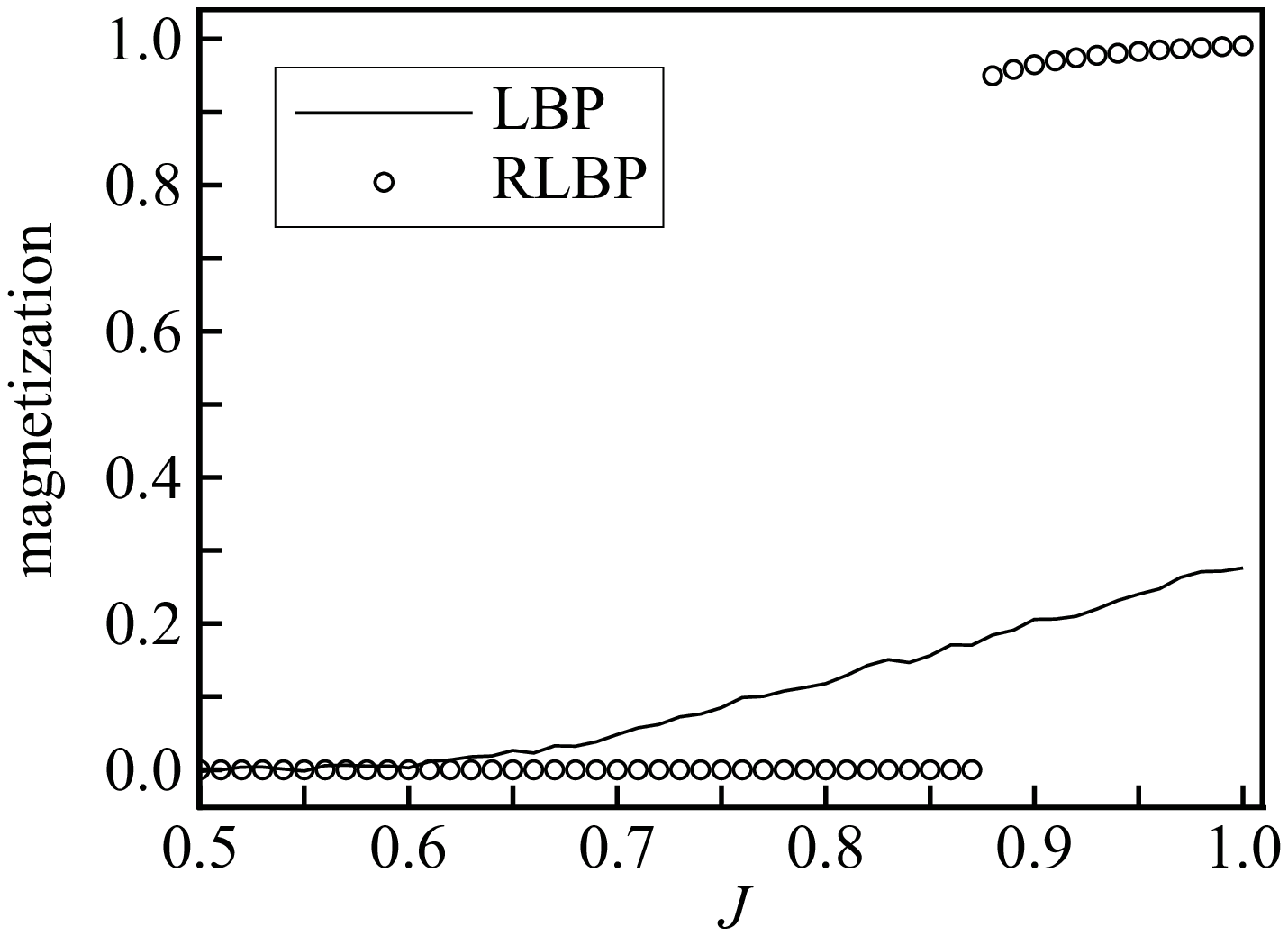}
\end{center}
\caption{Quenched magnetizations versus $J$, where $q = 2$ and $\sigma = 1$. }
\label{fig:criticality}
\end{figure}

Figure \ref{fig:criticality} shows the plot of the quenched average of the magnetizations, $M_{\mrm{LBP}}$ and $M_{\mrm{RLBP}}$, when the model shown in equation (\ref{eq:EA-Model}) is defined on a graph of a $14 \times 14$ square lattice with periodic boundary conditions and when $q = 2$. 
We observe that the two methods show the different nature of the magnetizations. 
The magnetization obtained by standard LBP continuously increases with the increase in $J$,  
whereas that obtained by the proposed method drastically increases, like a first-order transition, around $J \approx 0.88$.   
This different physical picture probably causes the mismatches between the two methods in the cases of large $J$ in figures \ref{fig:8grid-Q2}--\ref{fig:8grid-Q4}.

\begin{figure*}[htb]
\begin{center}
\includegraphics[height=4.5cm]{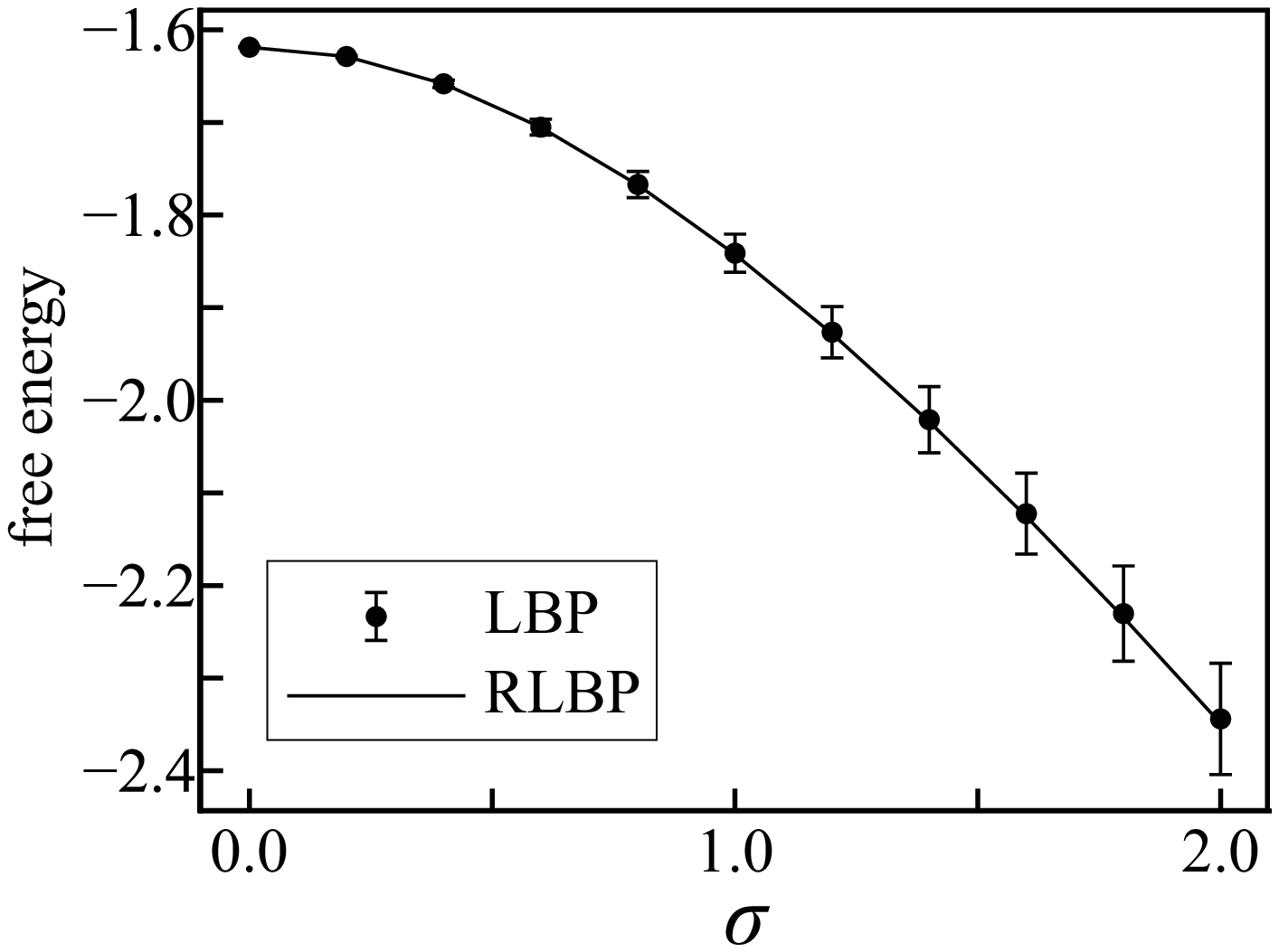}
\includegraphics[height=4.5cm]{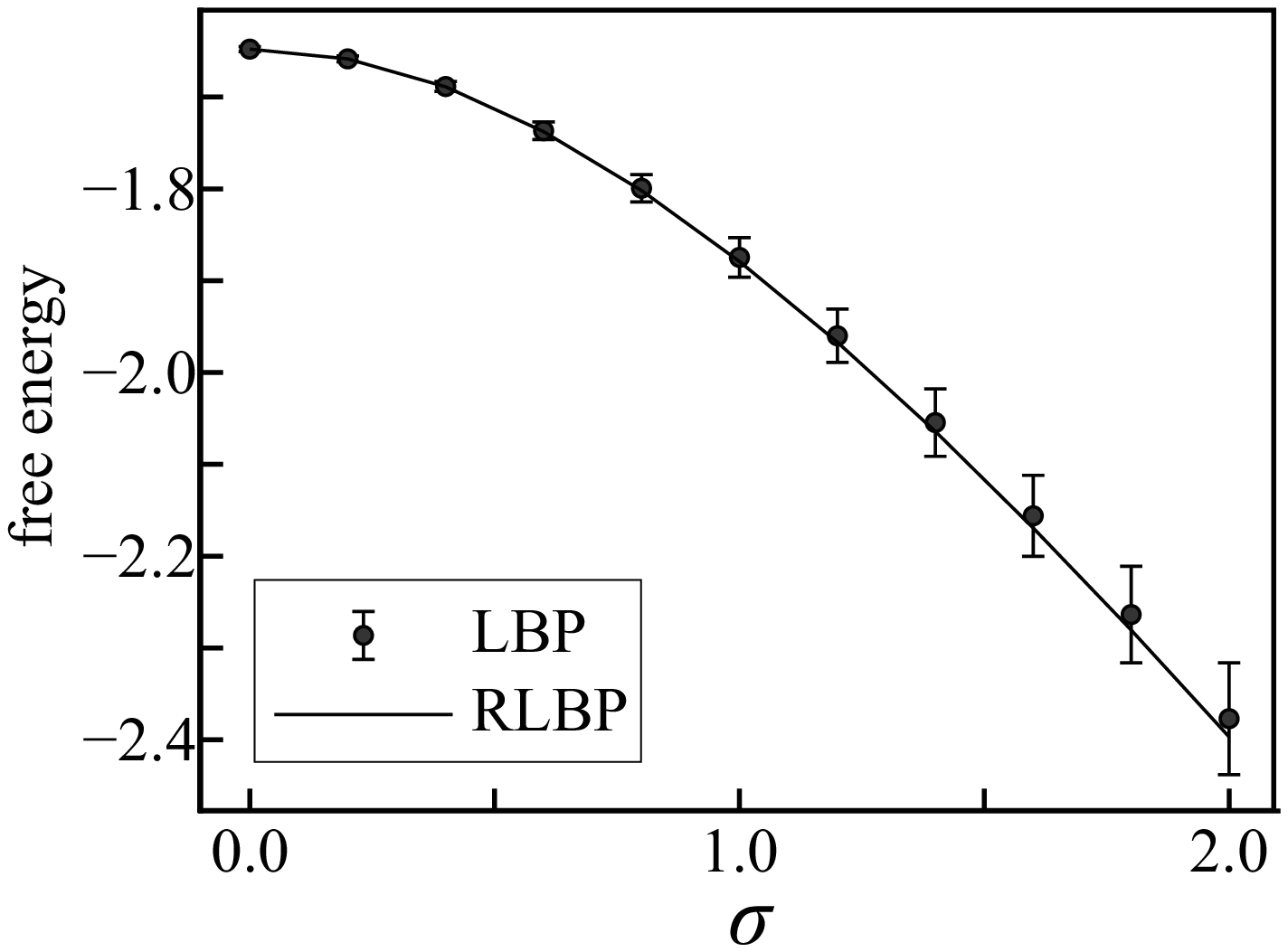}
\end{center}
\caption{Quenched Bethe free energies per variable versus $\sigma$ for $q = 5$. The left panel shows the free energies when $\delta = 0.2$, and the right panel shows the free energies when $\delta = 0.4$. The error bars are the standard deviation.}
\label{fig:Q5-disorder}
\end{figure*}

Our formulation allows for the approximate evaluation of the quenched average of Bethe free energy over the random fields with disordered interactions.
We show the results when the model in equation (\ref{eq:EA-Model}) is defined on a graph of a $14 \times 14$ square lattice with free boundary conditions 
and when the interactions $\{J_{ij}\}$ are independently drawn from $\mcal{N}(J_{ij} \mid 0, \delta^2)$. 
Figure \ref{fig:Q5-disorder} shows the results of the free energies versus $\sigma$ for $q = 5$. 
Each plot obtained by LBP is numerically averaged over 100 realizations of the random fields and over 200 realizations of the interactions, 
and that obtained by our method is averaged over 200 realizations of the interactions. 
Since the error bars of our method are quite small compared to LBP, we omit them in the figure. 
The results obtained by our method are consistent with the numerically averaged Bethe free energies.

\begin{figure}[htb]
\begin{center}
\includegraphics[height=4.5cm]{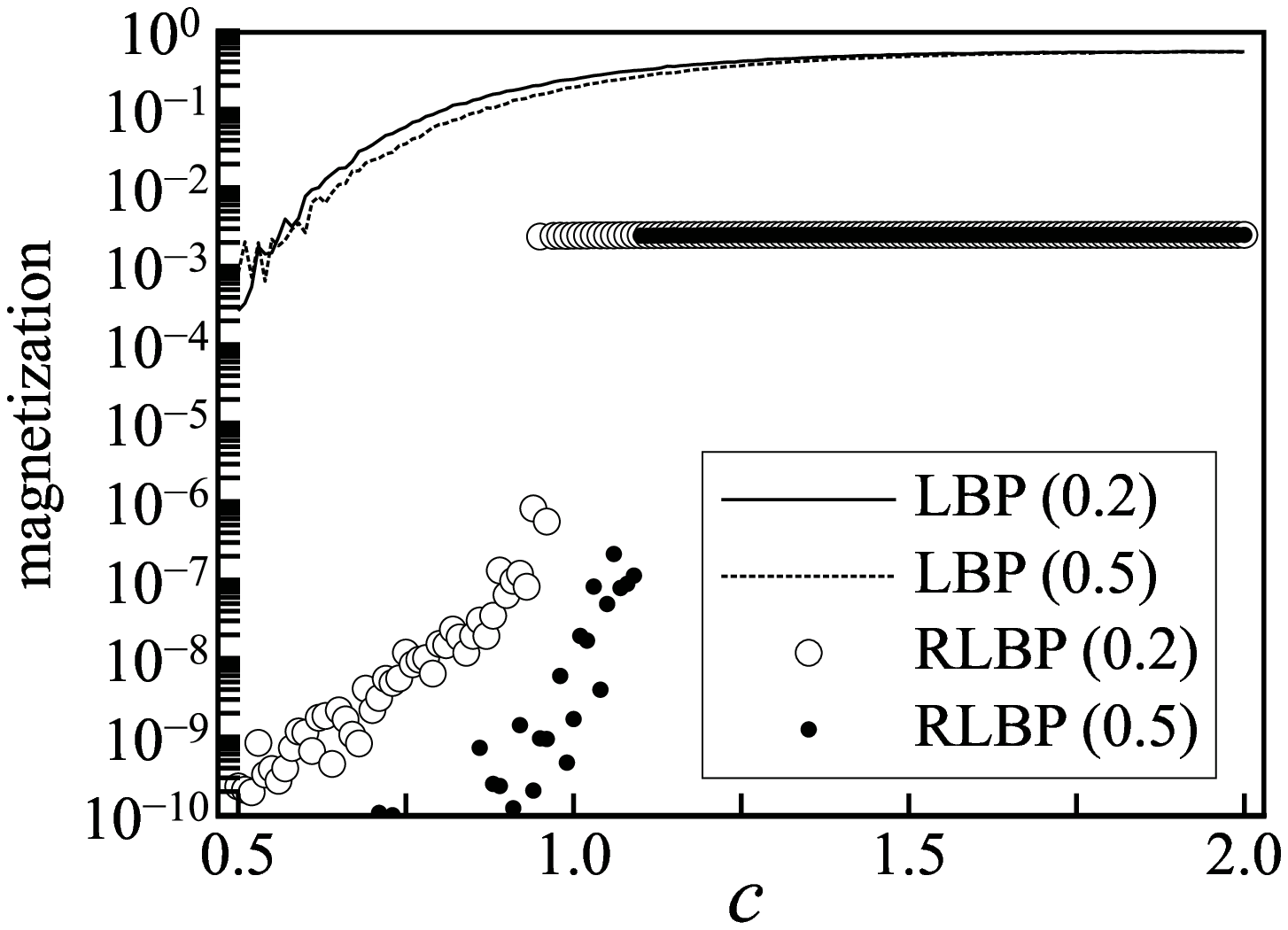}
\end{center}
\caption{Quenched magnetizations versus $c$ for $\delta = 0.2$ and $\delta = 0.5$ when $q = 2$ and $\sigma = 1$. }
\label{fig:criticality-square}
\end{figure}

To see the effect of the disorder in the interactions on the behavior of the magnetization, we show the plot of the quenched average of the magnetizations when the model shown in equation (\ref{eq:EA-Model}) is defined on a graph of a $14 \times 14$ square lattice with periodic boundary conditions and when $q = 2$ and the interactions $\{J_{ij}\}$ are independently drawn from $\mcal{N}(J_{ij} \mid c, \delta^2)$ in figure \ref{fig:criticality-square}. 
We observe that the magnetizations obtained by our method show the first-order transition, as in figure \ref{fig:criticality}. 
However, the values of the magnetizations after the transition are quite small compared to those in figure \ref{fig:criticality}.

\subsubsection{Random Regular Graph}
\label{sec:RRG}

\begin{figure}[htb]
\begin{center}
\includegraphics[height=4.5cm]{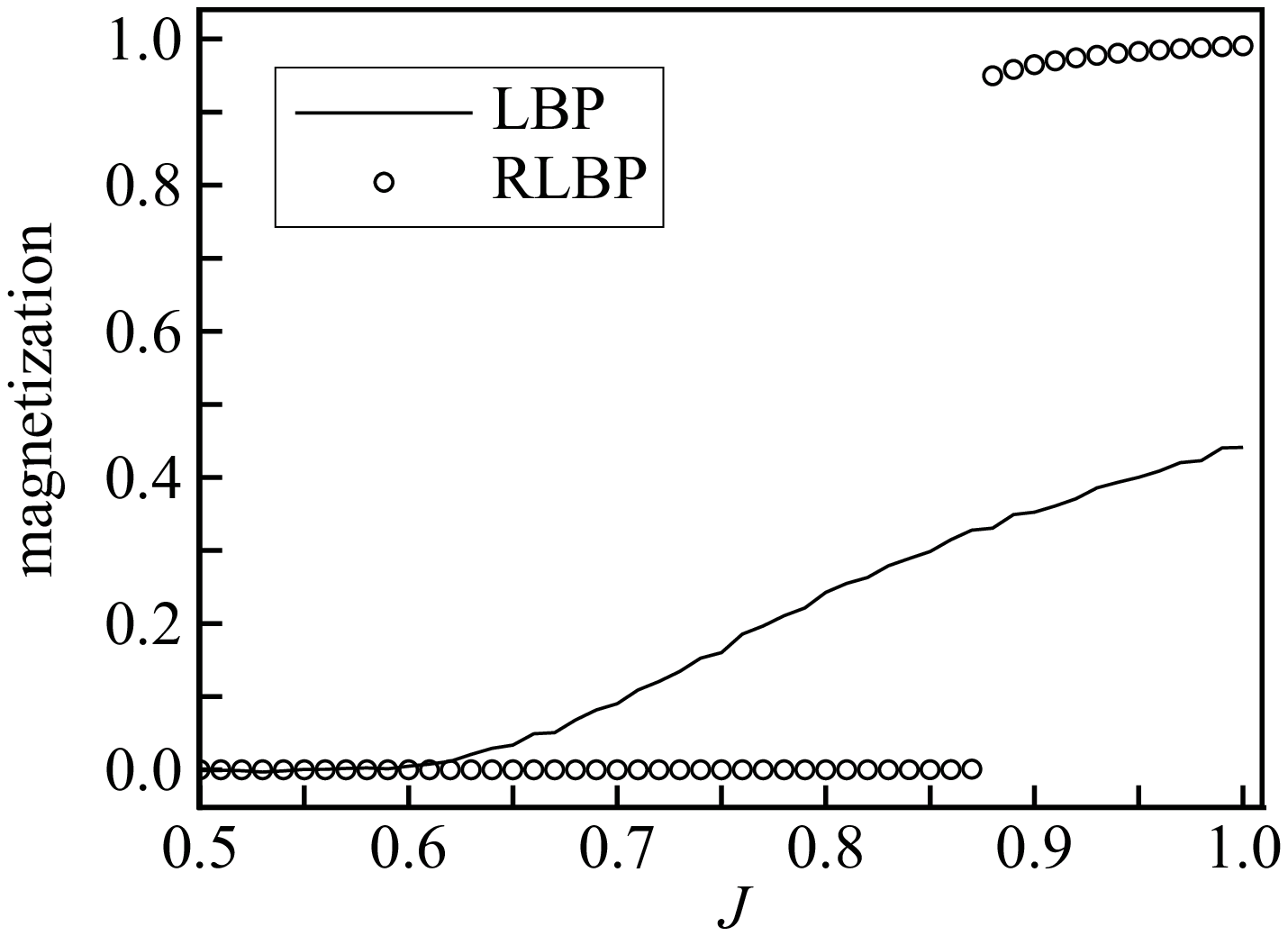}
\end{center}
\caption{Quenched magnetizations versus $J$ on the RRG, where $q = 2$ and $\sigma = 1$. }
\label{fig:criticality-RRG}
\end{figure}

A random regular graph (RRG) is a random graph in which the degrees of all vertices are fixed by the constant $d$.  
\begin{figure*}[htb]
\begin{center}
\includegraphics[height=4.5cm]{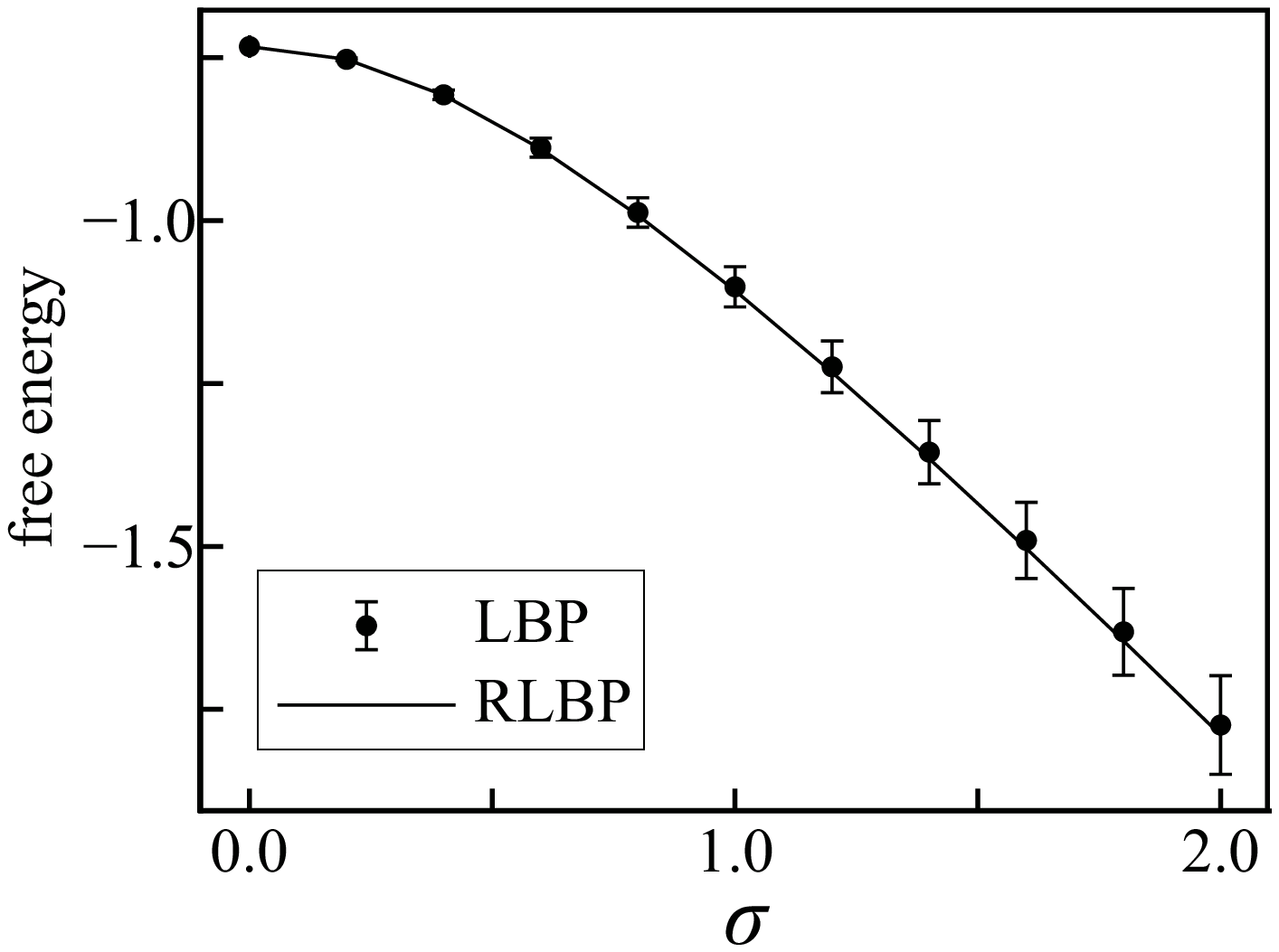}
\includegraphics[height=4.5cm]{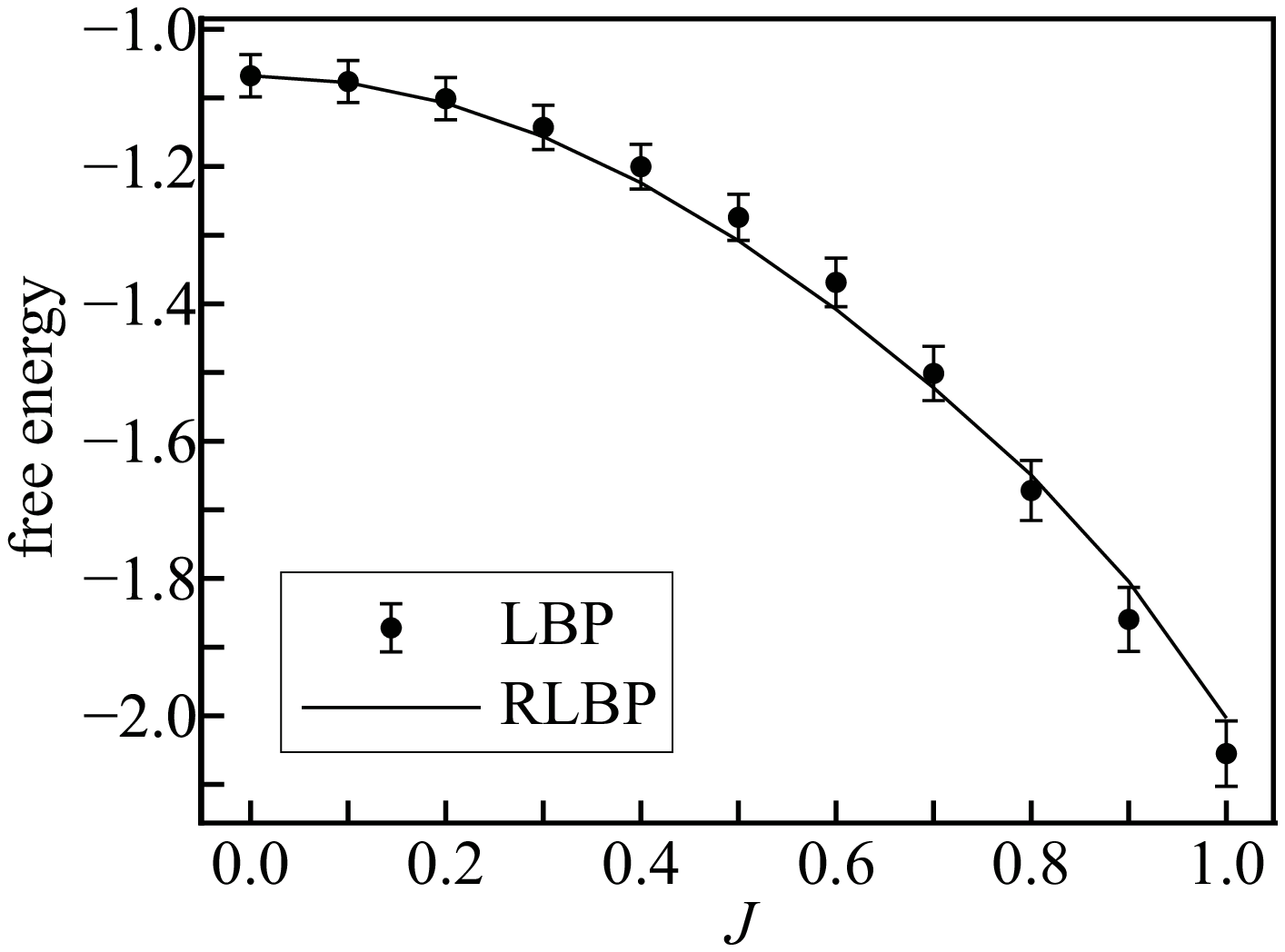}
\end{center}
\caption{Quenched Bethe free energies per variable for $q = 2$ on the RRG. The left panel shows the free energies versus $\sigma$ with $J = 0.2$, and the right panel shows the free energies versus $J$ with $\sigma = 1$. The error bars are the standard deviation.}
\label{fig:Q2-RRG4}
\end{figure*}
Figure \ref{fig:Q2-RRG4} shows the results when the model in equation (\ref{eq:EA-Model}) is defined on an RRG with 200 vertices and $d= 4$ 
and when $J_{ij} = J$ and $q = 2$.
Each plot obtained by LBP is numerically averaged over 100 realizations of the random fields and over 100 realizations of the structure of graph, 
and that obtained by our method is averaged over 100 realizations of the structure of graph. 
Since the error bars of our method are quite small compared to LBP, we omit them in the figure, as in figure \ref{fig:Q5-disorder}. 
The behaviors of the quenched magnetizations, $M_{\mrm{LBP}}$ and $M_{\mrm{RLBP}}$, obtained by the two methods in this case are shown in figure \ref{fig:criticality-RRG}.
The behaviors of the quenched magnetizations in this figure are similar to those shown in figure \ref{fig:criticality}.

LBP is asymptotically justified on an RRG~\cite{RFIM2010}, because an RRG is quite sparse. 
Therefore, we can expect that the results obtained by LBP are close to the exact solutions.
Except for the RS assumption, our method consists of two approximations: the approximation in equation (\ref{eq:CVM}) and the approximation in equation (\ref{eq:BetheApproximation-V^alpha}). 
Since the latter approximation is the Bethe approximation, it can be justified on a sparse graph such as an RRG. 
This suggests that the mismatch between the two methods in the right panel in figure \ref{fig:Q2-RRG4} is mainly caused by the first approximation, 
and that the first approximation produces the metastable state that causes the first-order transition in figure \ref{fig:criticality-RRG}.

\begin{figure*}[htb]
\begin{center}
\includegraphics[height=4.5cm]{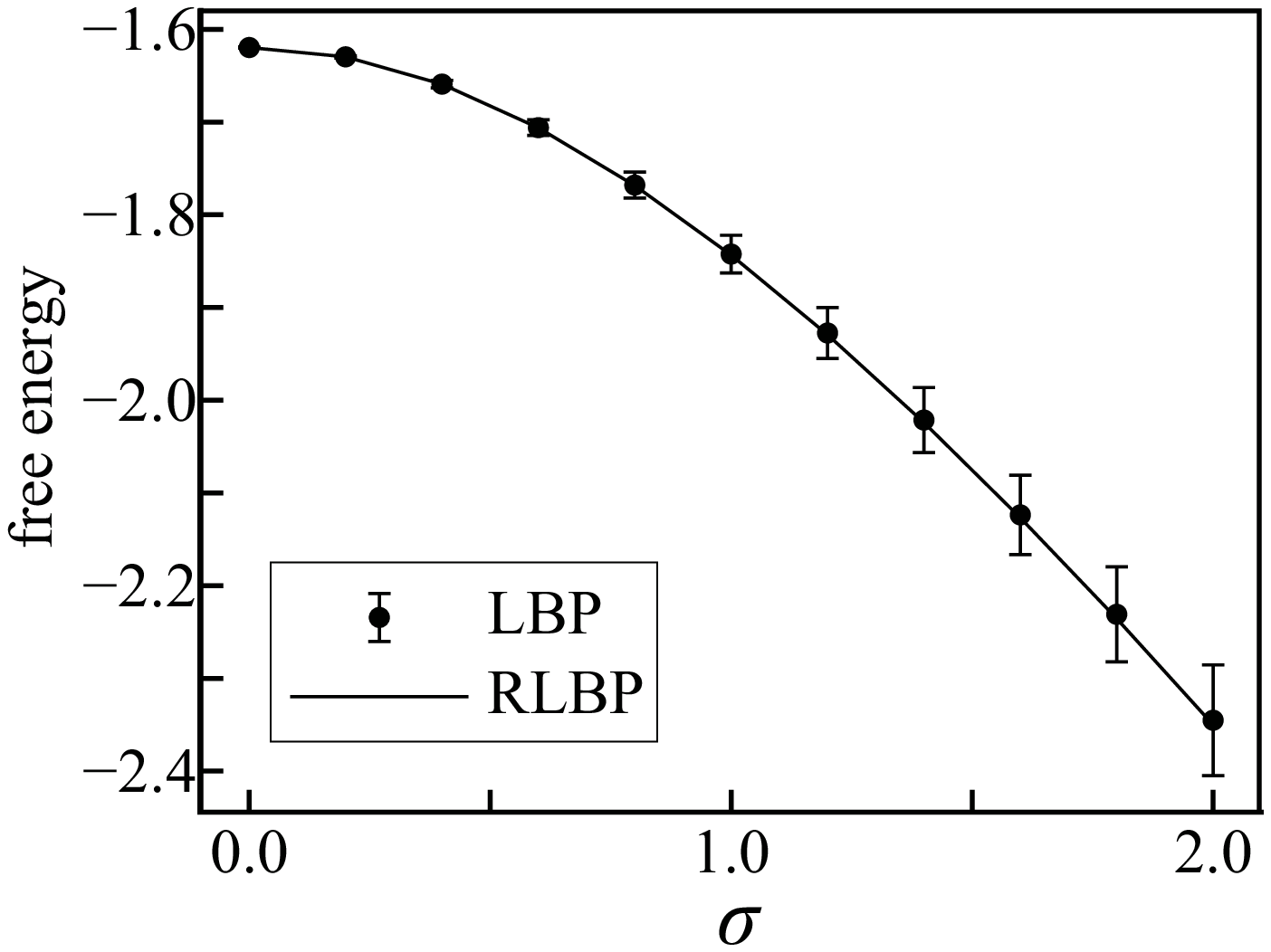}
\includegraphics[height=4.5cm]{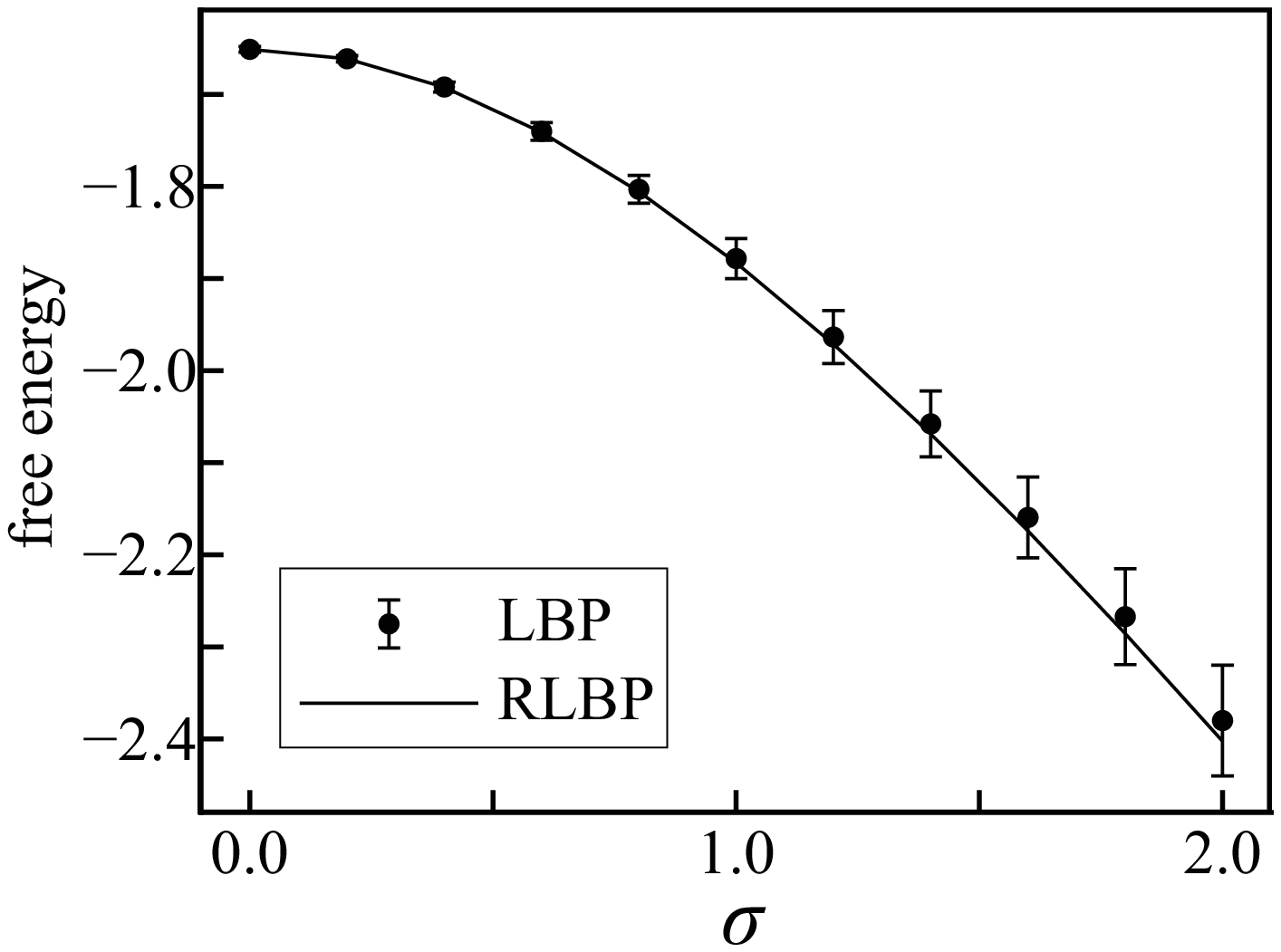}
\end{center}
\caption{Quenched Bethe free energies per variable versus $\sigma$ for $q = 5$ on the RRG. The left panel shows the free energies when $\delta = 0.2$, and the right panel shows the free energies when $\delta = 0.4$. The error bars are the standard deviation.}
\label{fig:Q5-RRG-disorder}
\end{figure*}

\begin{figure}[htb]
\begin{center}
\includegraphics[height=4.5cm]{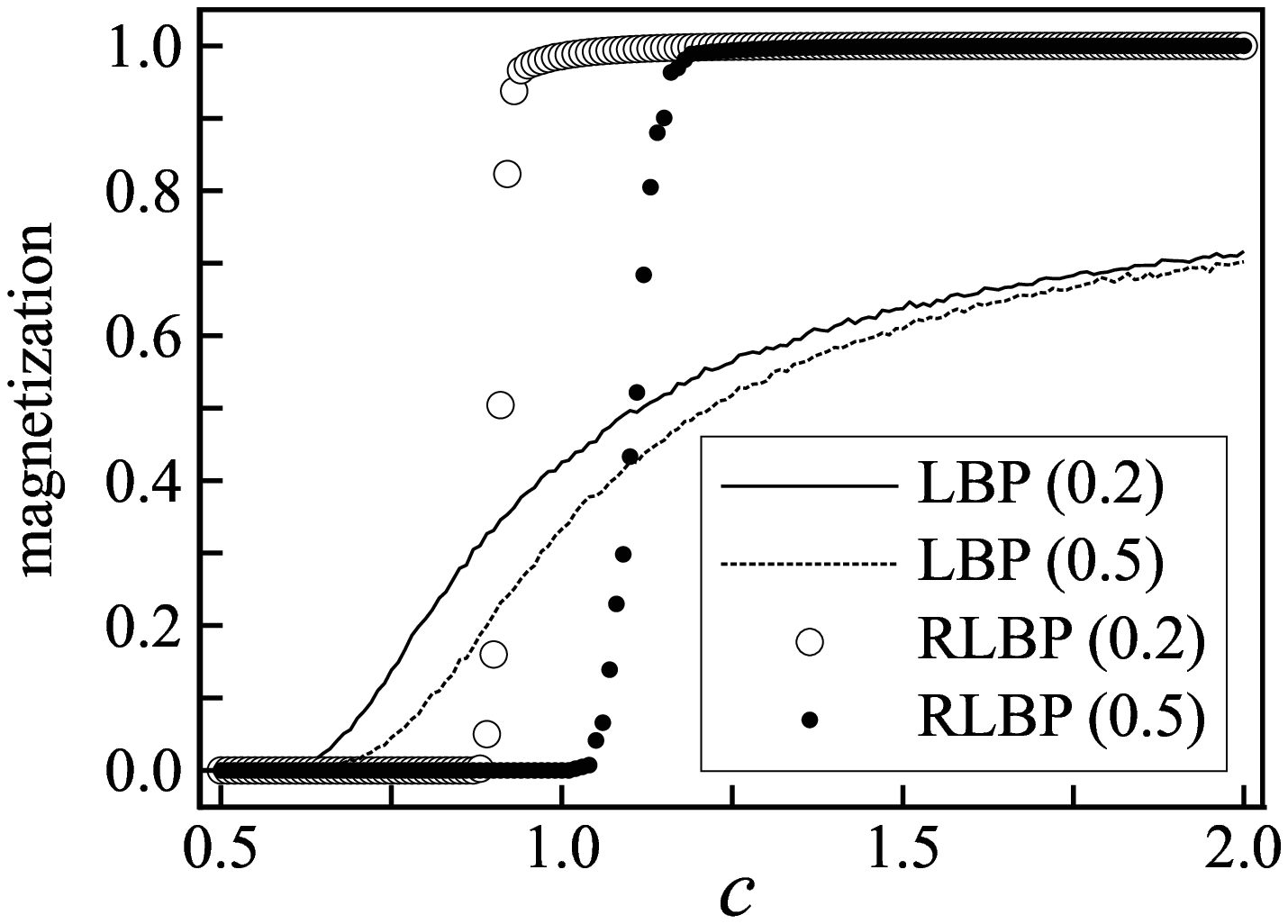}
\end{center}
\caption{Quenched magnetizations versus $c$ on the RRG for $\delta = 0.2$ and $\delta = 0.5$ when $q = 2$ and $\sigma = 1$. }
\label{fig:criticality-RRG-disorder}
\end{figure}

As in section \ref{sec:munerical-square}, we again see the case with the disordered interactions. 
Figure \ref{fig:Q5-RRG-disorder} shows the plots of the quenched Bethe free energies versus $\sigma$ for $q = 5$ 
when the model in equation (\ref{eq:EA-Model}) is defined on an RRG with 200 vertices and $d= 4$ 
and when the interactions $\{J_{ij}\}$ are independently drawn from $\mcal{N}(J_{ij} \mid 0, \delta^2)$. 
Each plot in the figure is obtained in the same manner as the case in figure \ref{fig:Q5-disorder}.
As is the case in figure \ref{fig:Q5-disorder}, the results of our method are consistent with the numerically averaged Bethe free energies.
Figure \ref{fig:criticality-RRG-disorder} shows the plot of the quenched average of the magnetizations when the model in equation (\ref{eq:EA-Model}) is defined on an RRG with 200 vertices and $d= 4$ 
and when $q = 2$ and $\{J_{ij}\}$ are independently drawn from $\mcal{N}(J_{ij} \mid c, \delta^2)$. 
It can be observed that the transition of the magnetization obtained by our method is nearly continuous with the increase in the magnitude of the disorder. 
This suggests that the disorder in the interactions violates the metastable state of the quenched Bethe free energy obtained by our method 
in the case of an RRG.

\subsection{Exactly Solvable Case -- Ferromagnetic Mean-field Model in Random Fields}
\label{sec:exact-case}

In this section, we consider the ferromagnetic mean-field model in random fields expressed as~\cite{RFIM1977}
\begin{align}
P(\bm{S}\mid \bm{h}, \beta)\propto \exp \beta \Big( \sum_{i \in V}\phi(S_i, h_i) + \frac{1}{n}\sum_{i<j}g(S_i)g(S_j)\Big),
\label{eq:FerroMeanFieldModel}
\end{align}
where the second summation represents the summation over all of the distinct pairs of vertices, and $\{h_i\}$ represents the i.i.d. random fields drawn from the distribution $p_h(h_i)$.
By using the Hubbard--Stratonovich transformation, the free energy (per variable) of this model, $f := F(\bm{h}, \beta) / n$, can be expressed as
\begin{align*}
f&= - \frac{1}{n\beta}\ln \int \diff m \exp n \Big\{ - \frac{\beta}{2}m^2 \nn
\aleq
+ \frac{1}{n}\sum_{i \in V}\ln \sum_{S} 
\exp \beta \Big(\phi(S, h_i) + m g(S) - \frac{g(S)^2}{2n}\Big)\Big\} \nn
\aleq
-\frac{1}{2n \beta}\ln \frac{n \beta}{2\pi}.
\end{align*}
For a sufficiently large $n$, we can compute the integration by using the saddle point method and the summation over $i \in V$ in the exponent by using the law of large numbers, 
so that the free energy can be expressed as
\begin{align}
f&= \frac{\beta}{2}m^2 - \frac{1}{\beta}\int \diff h\, p_h(h)\ln \sum_{S} \exp \beta \big(\phi(S, h) + m g(S) \big) 
\label{eq:FreeEnergy_FerroMeanFieldModel}
\end{align}
in the thermodynamic limit, where $m$ is the solution to the saddle point equation:
\begin{align}
m = \int \diff h\, p_h(h)\frac{\sum_{S} g(S) \exp \beta \big(\phi(S, h) + m g(S) \big) }{\sum_{S} \exp \beta \big(\phi(S, h) + m g(S) \big) }.
\label{eq:Magnetization_FerroMeanFieldModel}
\end{align}
Since the free energy in equation (\ref{eq:FreeEnergy_FerroMeanFieldModel}) does not depend on $\{h_i\}$, 
the quenched average of the free energy over the random fields, $[F]_{\bm{h}}/n$, coincides with equation (\ref{eq:FreeEnergy_FerroMeanFieldModel}).

Since $\psi_{i,j}(S_i,S_j) = g(S_i)g(S_j) / n$, the message-passing equation in equation (\ref{eq:MessagePassing-LBP(RS)}) can be expanded as
\begin{align}
&\ln \mu_{j \to i}(S_i) \nn
&= \ln \sum_{S_j}Q_j(S_j)\exp \big( \beta\psi_{i,j}(S_i,S_j)\big)\mu_{i \to j}(S_j)^{-1} + c_0\nn
&=\frac{\beta}{n}\frac{g(S_i)\sum_{S_j}g(S_j)Q_j(S_j)\mu_{i \to j}(S_j)^{-1}}{\sum_{S_j}Q_j(S_j)\mu_{i \to j}(S_j)^{-1}}
+c_1 +O(n^{-2}),
\label{eq:MessagePassing-LBP(RS)_FerroMeanFieldModel}
\end{align}
where $c_0$ and $c_1$ are constants unrelated to $S_i$.
From this equation, we ensure that all of the messages are constants unrelated to $\bm{S}$ for a sufficiently large $n$, because $\mu_{j \to i}(S_i) = \exp(c_1 + O(n^{-1})) \simeq  \exp(c_1)$. 
Therefore, from equations (\ref{eq:exp(Lambda)}) and (\ref{eq:MessagePassing-LBP(RS)_FerroMeanFieldModel}), we obtain
\begin{align}
\Lambda_i(S_i) = \frac{g(S_i)}{n}\sum_{j \in \partial i}\sum_{S_j}g(S_j)Q_j(S_j) + c_2
=m g(S_i) + c_2
\label{eq:exp(Lambda)_FerroMeanFieldModel}
\end{align}
for a sufficiently large $n$, where $c_2$ is a constant unrelated to $S_i$, and we redefine $m:= n^{-1}\sum_{i \in V}\sum_{S_i}g(S_i)Q_j(S_i)$.
By substituting equation (\ref{eq:exp(Lambda)_FerroMeanFieldModel}) into equation (\ref{eq:Qi-LBP(RS)}), we obtain the same expression for $m$ as equation (\ref{eq:Magnetization_FerroMeanFieldModel}).
Since all of the messages are constants, from equation (\ref{eq:Qij-LBP(RS)}), we have
\begin{align*}
Q_{i,j}(S_i,S_j) = Q_i(S_i)Q_j(S_j).
\end{align*}
By substituting this equation and equation (\ref{eq:exp(Lambda)_FerroMeanFieldModel}) into equation (\ref{eq:quenched-FreeEnergy-Bethe(RS)}), we find that the equality
$
f = \mcal{F}_{\mrm{min}}^{\mrm{LBP(RS)}} / n
$
holds in the thermodynamic limit. 
From this result, we can conclude that our message-passing method can exactly compute the quenched average of the free energy of the model in equation (\ref{eq:FerroMeanFieldModel}) in the thermodynamic limit.

\section{Application to Bayesian Image Restoration}\label{sec:ImageRestoration}

In this section, we describe the estimation of the statistical performance of the Bayesian image restoration system using LBP by using the proposed method. 
In images, each pixel is allocated in a two-dimensional grid and has an intensity value corresponding to the color at its position. 
For an image $\bm{I} \in \{I_i \mid i \in V\}$, the entry $I_i \in \{0,1,\ldots,q-1\}$ represents the intensity of the $i$th pixel. 

\begin{figure}[htb]
\begin{center}
\includegraphics[height=3.0cm]{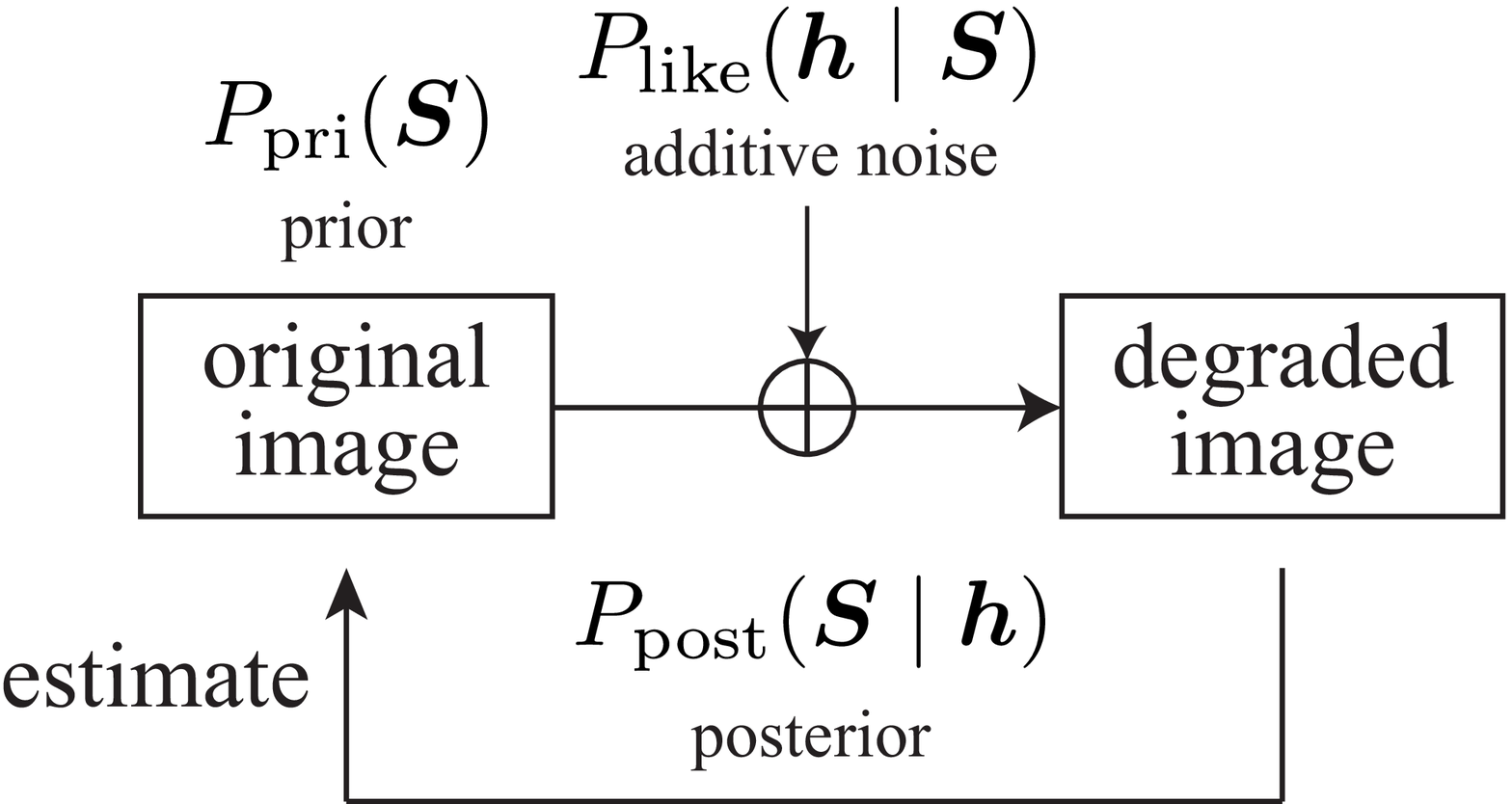}
\end{center}
\caption{Scheme of Bayesian image restoration.}
\label{fig:SchemeBayes}
\end{figure}

Suppose that the original image $\bm{I}$ is degraded to $\bm{h}$ through a specific noise process.
In the Bayesian image restoration system, we assume that the original image is the sample drawn from the specific distribution $P_{\mrm{pri}}(\bm{S})$ known as \textit{the prior distribution}. 
We observe only the degraded image as the input, and we want to estimate the original image from the input degraded image. 
To do the image restoration, we compute \textit{the posterior distribution} of the original image, 
$P_{\mrm{post}}(\bm{S} \mid \bm{h}) \propto P_{\mrm{like}}(\bm{h} \mid \bm{S})P_{\mrm{pri}}(\bm{S})$, 
where $P_{\mrm{like}}(\bm{h} \mid \bm{S})$ is the noise process referred to as \textit{the likelihood}.
We use the posterior distribution to produce the restored image as the output.
The scheme of Bayesian image restoration is shown in figure \ref{fig:SchemeBayes}.

\subsection{Bayesian Image Restoration using LBP}
\label{sec:ImageRestoration-LBP}

For the original image $\bm{I}$, suppose that the degraded image $\bm{h}$ is generated by adding additive white Gaussian noise, i.e., $h_i = I_i + \eta_i$, where $\eta_i$ is the random noise drawn from  
the Gaussian distribution $\mcal{N}(\eta \mid 0, \sigma^2)$.

We define the prior distribution of the images $\bm{S}$ as
\begin{align}
P_{\mrm{pri}}(\bm{S} \mid \alpha) \propto \exp\Big(\alpha \sum_{\{i,j\} \in E} \xi(S_i,S_j)\Big),
\label{eq:prior}
\end{align} 
which is defined on a square lattice corresponding to the configuration of pixels.
The energy function $\xi(S_i,S_j)$ defines the relationship among neighboring pixels, and it often takes a form that emphasizes the smoothness among neighboring pixels, 
e.g., $\xi(S_i,S_j) = - |S_i - S_i|$.
The positive parameter $\alpha$ controls the strength of the smoothness.
For the original image, the distribution of the degraded image, i.e., the likelihood, is expressed as
\begin{align}
P_{\mrm{like}}(\bm{h} \mid \bm{S} = \bm{I}, \sigma^2) = \prod_{i \in V}\mcal{N}(h_i \mid I_i, \sigma^2).
\label{eq:AWGN}
\end{align}
From equations (\ref{eq:prior}) and (\ref{eq:AWGN}), given a specific degraded image $\bm{h}$, the posterior distribution of the original image is obtained by
\begin{align}
&P_{\mrm{post}}(\bm{S} \mid \bm{h}, \alpha, \sigma^2)\propto P_{\mrm{like}}(\bm{h} \mid \bm{S}, \sigma^2)P_{\mrm{pri}}(\bm{S} \mid \alpha) \nn
&\propto
\exp\Big(-\sum_{i \in V}\frac{(S_i - h_i)^2}{2\sigma^2} + \alpha\sum_{\{i,j\} \in E} \xi(S_i,S_j)\Big).
\label{eq:posterior}
\end{align}
The posterior distribution is the special case of equation (\ref{eq:MRF}).
The degraded image is regarded as the random fields in the posterior distribution.

In maximum posterior marginal (MPM) estimation, the $i$th pixel of the restored image is obtained by 
\begin{align}
S_i^{\mrm{MPM}} := \argmax_{S_i} P_{\mrm{post}}(S_i \mid \bm{h}, \alpha, \sigma^2),
\label{eq:MPM}
\end{align}
where $P_{\mrm{post}}(S_i \mid \bm{h}, \alpha, \sigma^2)$ is the marginal distribution of the posterior distribution in equation (\ref{eq:posterior})~\cite{TanakaReview2002}.
In practice, we approximate the true marginal distributions by the beliefs obtained by LBP for the posterior distribution,
\begin{align}
S_i^{\mrm{MPM}} \approx S_i^{\mrm{MPM(LBP)}} := \argmax_{S_i} b_i(S_i),
\label{eq:MPM-LBP}
\end{align}
where the belief $b_i(S_i)$ is obtained by LBP described in section \ref{sec:LBP-detail} 
with $\phi_i(S_i, h_i) = -(S_i - h_i)^2 / 2\sigma^2$, $\psi_{i,j}(S_i,S_j) = \alpha \xi(S_i,S_j)$, and $\beta = 1$.
The performance of the restoration is often measured by the mean square error (MSE) between the original image and the restored image, 
\begin{align}
D(\bm{I}, \bm{h}, \alpha , \sigma^2) := \frac{1}{n}\sum_{i \in V}\big(I_i - S_i^{\mrm{MPM(LBP)}}\big)^2.
\label{eq:MSE}
\end{align}
This MSE is for the specific input degraded image that is drawn from the likelihood in equation (\ref{eq:AWGN}), 
and it should take different values for different degraded images in general. 

For the original image $\bm{I}$, we attempt to estimate the average value of the MSE in equation (\ref{eq:MSE}) over all possible degraded images,
\begin{align}
D_{\mrm{av}}(\bm{I}, \alpha , \sigma^2) := \big[ D(\bm{I}, \bm{h}, \alpha , \sigma^2)\big]_{\bm{h}},
\label{eq:av-MSE}
\end{align}
where $p_i(h) = \mcal{N}(h \mid I_i, \sigma^2)$, 
because the average value corresponds to the statistical performance of the presented Bayesian restoration system for the original image.
By considering the input degraded image as the quenched parameter, 
the right-hand side of equation (\ref{eq:av-MSE}) can be regarded as the quenched average of the MSE obtained by LBP.
Thus, we approximate it using the message-passing method presented in section \ref{sec:ReplicaMessagePassing}. 

Using the proposed message-passing method, we approximate equation (\ref{eq:av-MSE}) as
\begin{align}
D_{\mrm{av}}(\bm{I}, \alpha , \sigma^2) \approx \frac{1}{n}\sum_{i \in V}\int \diff h\,\mcal{N}(h \mid I_i, \sigma^2)\big(I_i - r_i(h)\big)^2, 
\label{eq:av-MSE_app}
\end{align}
where
\begin{align*}
r_i(h)=\argmax_{S}\big( \phi_i(S,h) + \Lambda_i(S) \big).
\end{align*}
The detailed derivation of this approximation is shown in appendix \ref{app:sec:Deriv-D_av}.

\subsection{Numerical Experiment}
\label{sec:Numerical-Bayes}

\begin{figure}[htb]
\begin{center}
\includegraphics[height=3.0cm]{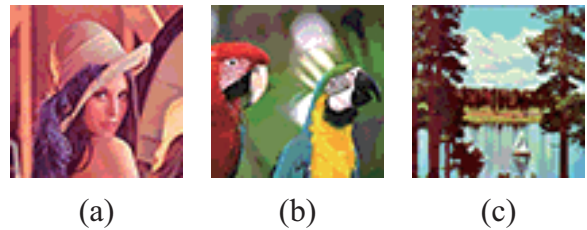}
\end{center}
\caption{Original 3-bit colored images $\bm{I}$: (a) lenna, (b) parrots, and (c) sailboat.}
\label{fig:original}
\end{figure}

In this section, we describe the estimation of the performance of Bayesian image restoration for the $64 \times 64$ original colored images shown in figure \ref{fig:original}.
Colored images consist of three different channels: red, green, and blue (RGB) channels, $\bm{I} = \{\bm{I}_{\mrm{red}} ,\bm{I}_{\mrm{green}}, \bm{I}_{\mrm{blue}}\}$.
The pixels in each channel in the original images have eight intensities, i.e., $q = 8$. 
In the following experiments, we assume that the three different channels are independently degraded by the same noise process in equation (\ref{eq:AWGN}), 
and we restore the generated degraded images, $\bm{h} = \{\bm{h}_{\mrm{red}} ,\bm{h}_{\mrm{green}}, \bm{h}_{\mrm{blue}}\}$, 
by separately applying the Bayesian image restoration based on the posterior distribution in equation (\ref{eq:posterior}) to the RGB channels. 
The value of $\sigma^2$ used in the restoration is the same as that used in the noise process, 
and we use the same values of the parameters, $\alpha$ and $\sigma^2$, in the restorations for the three different channels.
The total MSE of the restoration is obtained by the average of the MSEs of the RGB channels, i.e., 
$D(\bm{I}, \bm{h}, \alpha , \sigma^2)=\{D(\bm{I}_{\mrm{red}}, \bm{h}_{\mrm{red}}, \alpha , \sigma^2)+ D(\bm{I}_{\mrm{green}}, \bm{h}_{\mrm{green}}, \alpha , \sigma^2) 
+ D(\bm{I}_{\mrm{blue}}, \bm{h}_{\mrm{blue}}, \alpha , \sigma^2)\}/3$.

For the original colored images in figure \ref{fig:original}, we evaluated the average of the MSE, $D_{\mrm{av}}(\bm{I}, \alpha , \sigma^2)$, by using two types of methods: LBP and the proposed analytical method in equation (\ref{eq:av-MSE_app}). 
In LBP, we approximated $D_{\mrm{av}}(\bm{I}, \alpha , \sigma^2)$ by the sample average of $D(\bm{I}, \bm{h}, \alpha , \sigma^2)$ over 10000 different degraded images, which are generated from the original image $\bm{I}$ through the noise process shown in equation (\ref{eq:AWGN}).

\begin{figure*}[htb]
\begin{center}
\includegraphics[height=4.2cm]{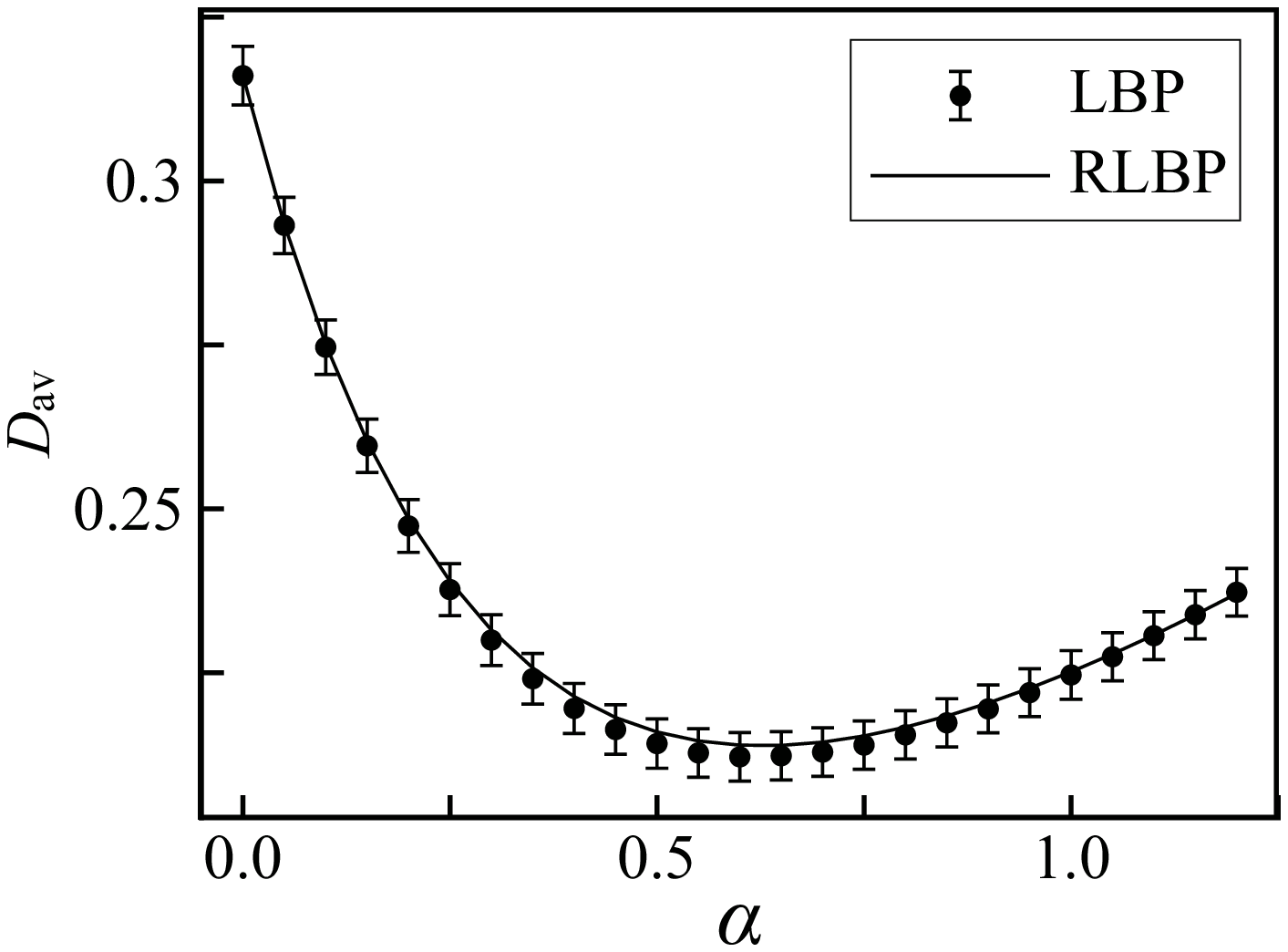}
\includegraphics[height=4.2cm]{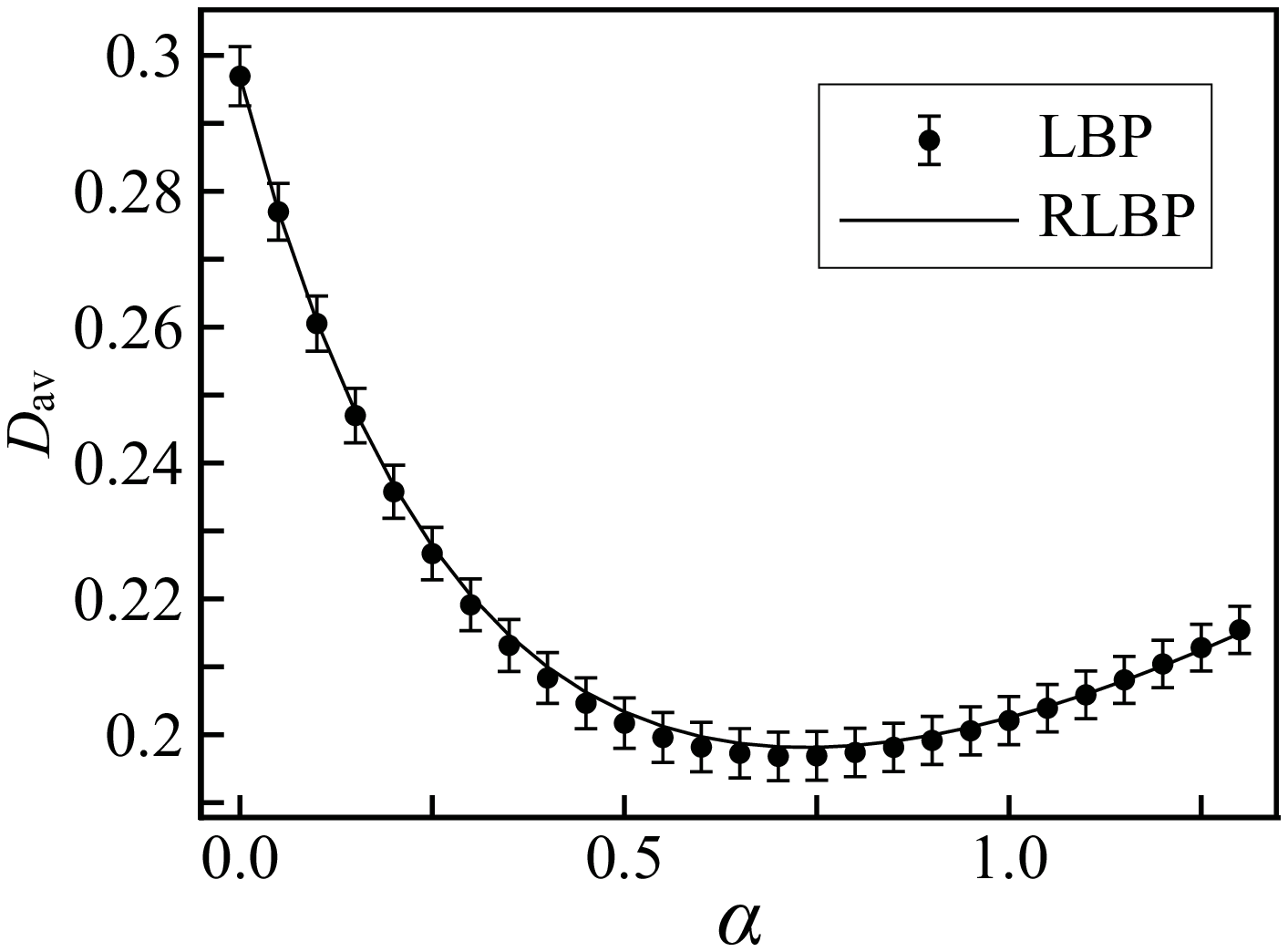}
\includegraphics[height=4.2cm]{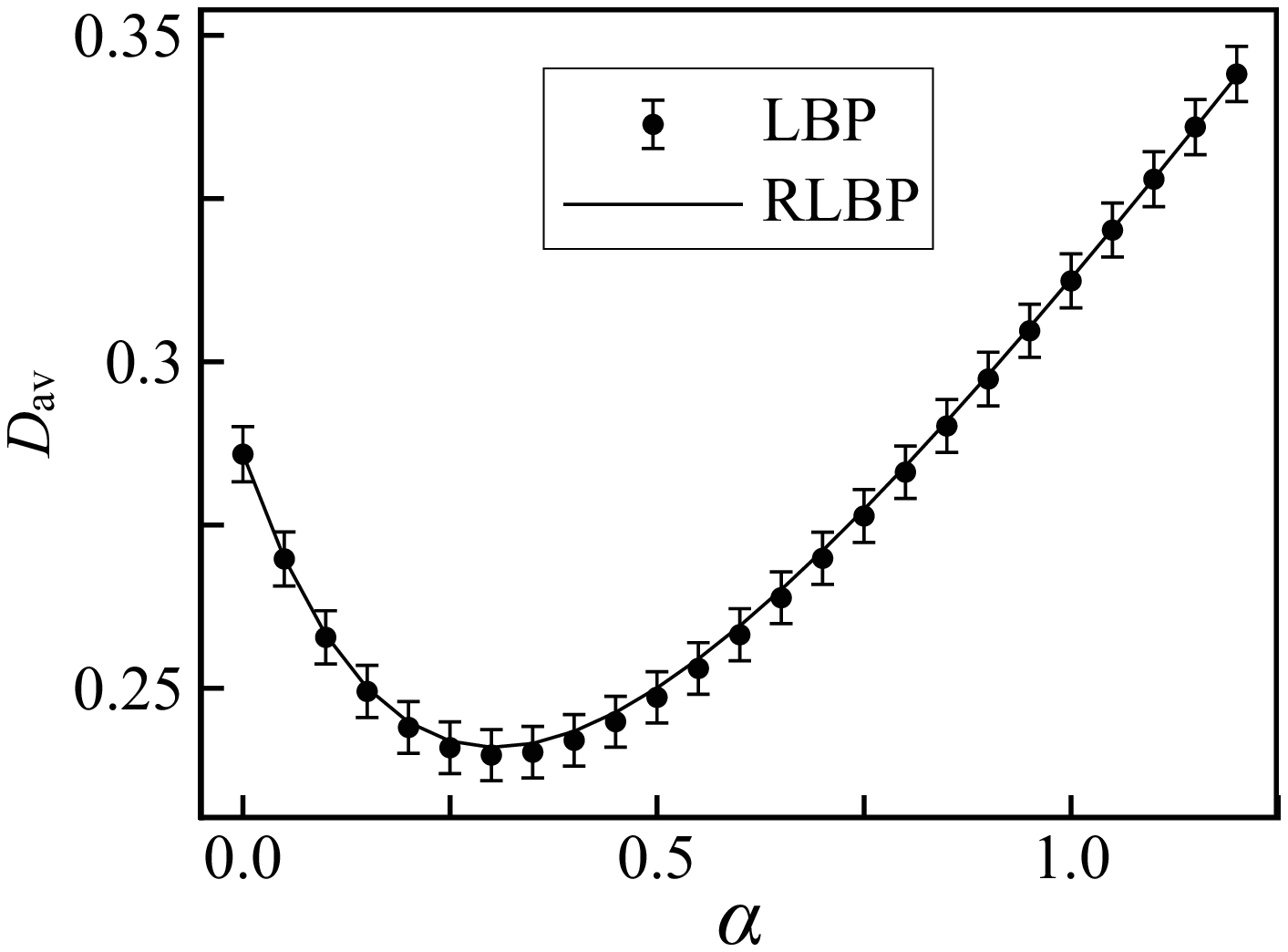}
\end{center}
\caption{Plots of $D_{\mrm{av}}(\bm{I}, \alpha , \sigma^2)$ versus $\alpha$ when $\sigma = 0.5$ and $\xi(S_i, S_j) = -(S_i - S_j)^2 / 2$ 
for the original images shown in figure \ref{fig:original}: the left, middle, and right plots show the results for figures \ref{fig:original}(a), \ref{fig:original}(b), and \ref{fig:original}(c), respectively. 
The error bars are the standard deviation.}
\label{fig:result_al}
\end{figure*}

\begin{figure*}[htb]
\begin{center}
\includegraphics[height=4.2cm]{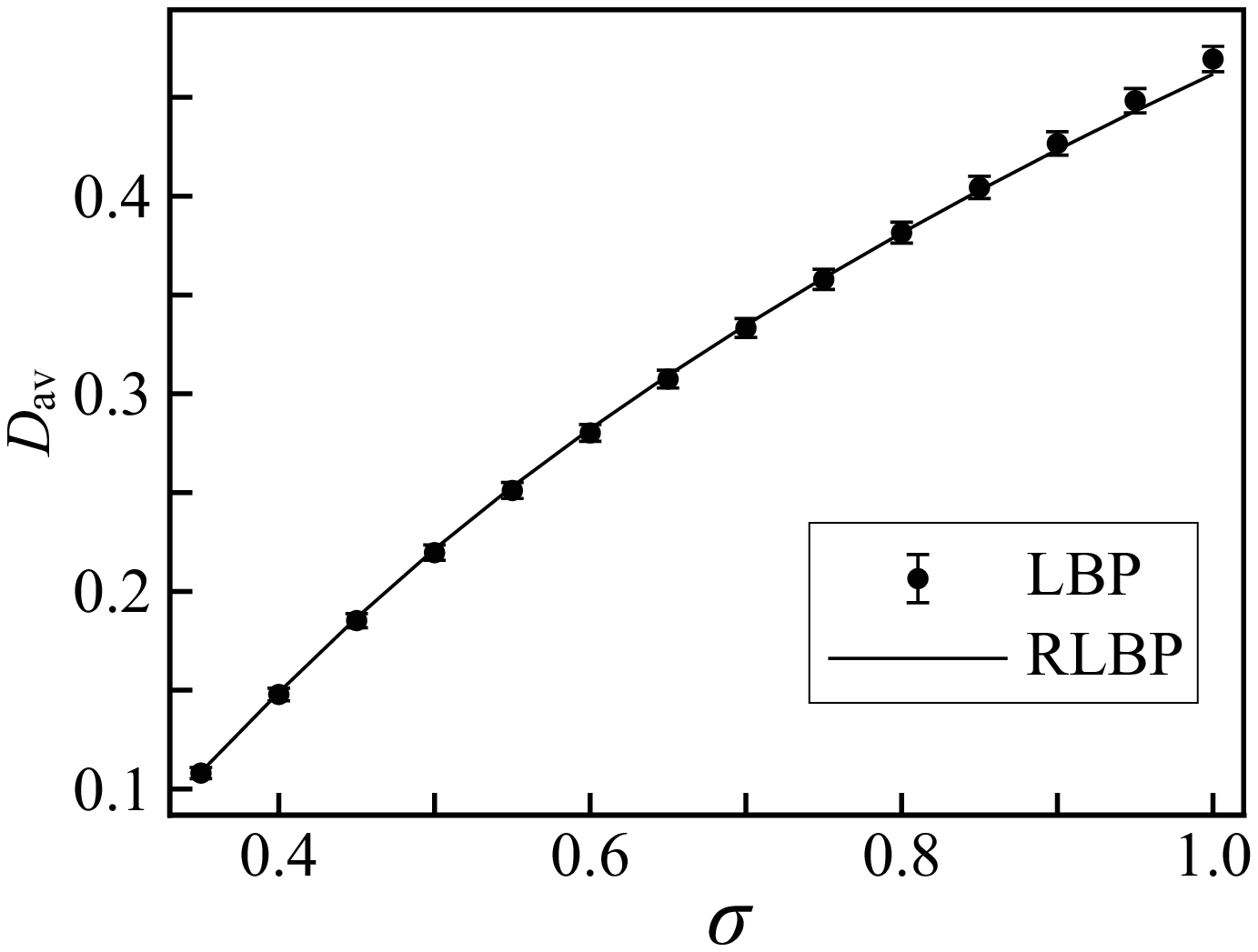}
\includegraphics[height=4.2cm]{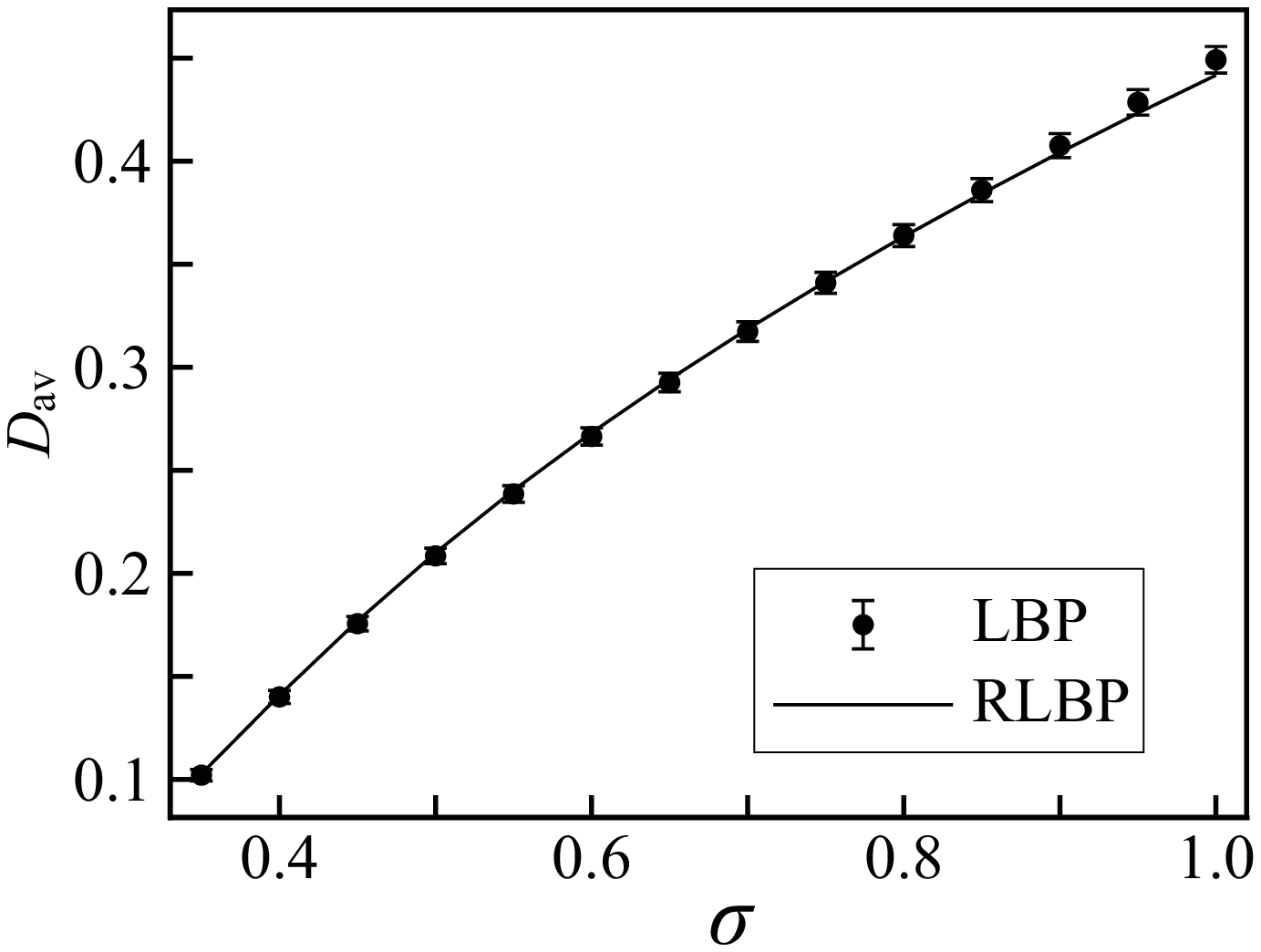}
\includegraphics[height=4.2cm]{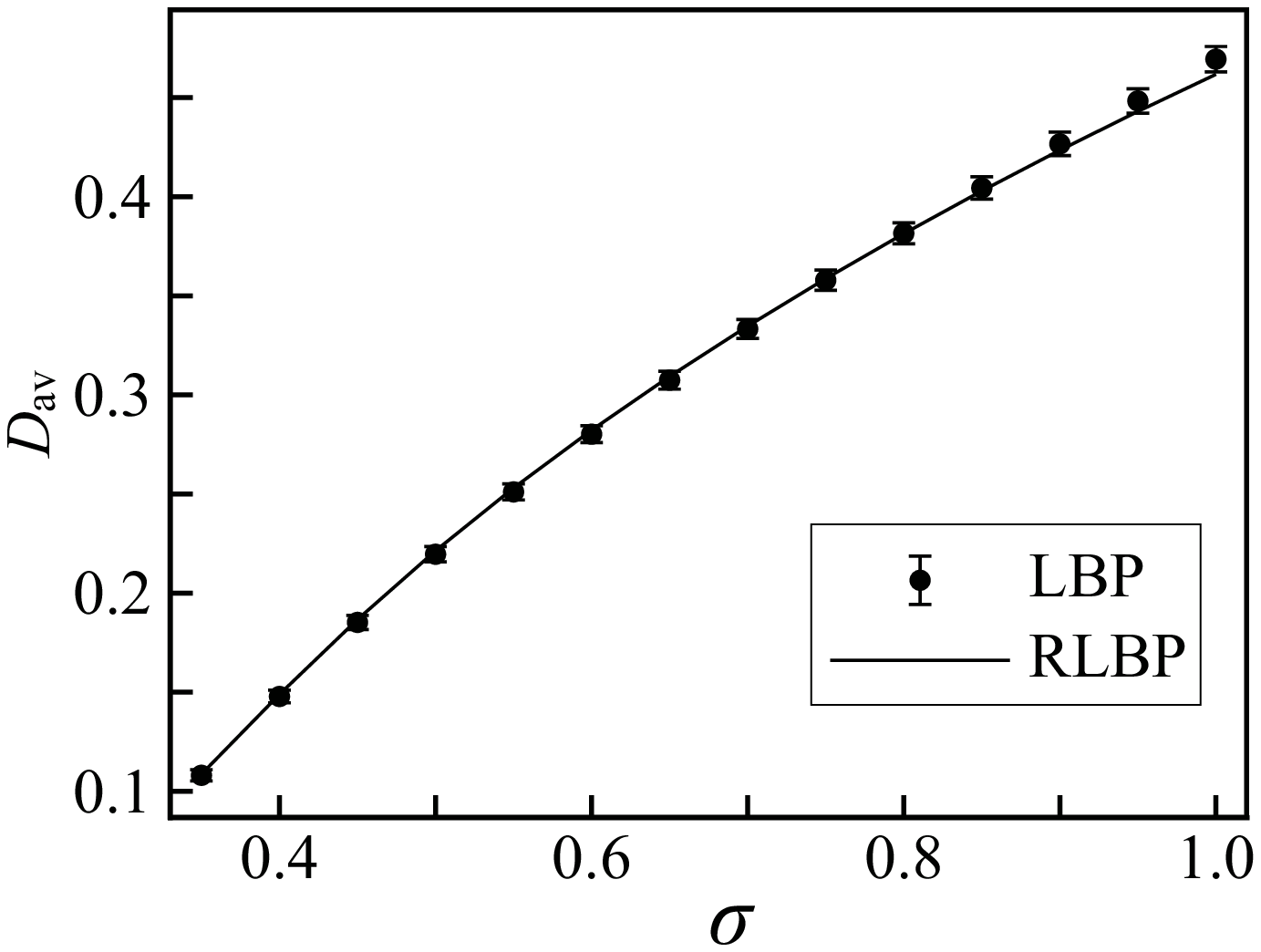}
\end{center}
\caption{Plots of $D_{\mrm{av}}(\bm{I}, \alpha , \sigma^2)$ versus $\sigma$ when $\alpha = 0.4$ and $\xi(S_i, S_j) = -(S_i - S_j)^2 / 2$ 
for the original images shown in figure \ref{fig:original}: the left, middle and right plots show the results for figures \ref{fig:original}(a), \ref{fig:original}(b), and \ref{fig:original}(c), respectively. 
The error bars are the standard deviation.}
\label{fig:result_sig}
\end{figure*}

At first, we show the results obtained by setting the function $\xi(S_i, S_j)$ as $\xi(S_i, S_j) = -(S_i - S_j)^2 / 2$ in the prior distribution in equation (\ref{eq:prior}).  
In figures \ref{fig:result_al} and \ref{fig:result_sig}, we show the plots of $D_{\mrm{av}}(\bm{I}, \alpha , \sigma^2)$ versus $\alpha$ and $\sigma$, respectively, 
obtained by the two different methods: the sample average of the LBP restorations, ``LBP,'' and the proposed analytical method, ``RLBP.''
Figure \ref{fig:result_al} shows the plots of $D_{\mrm{av}}(\bm{I}, \alpha , \sigma^2)$ versus $\alpha$ when $\sigma = 0.5$, 
and figure \ref{fig:result_sig} shows the plots of $D_{\mrm{av}}(\bm{I}, \alpha , \sigma^2)$ versus $\sigma$ when $\alpha = 0.4$. 
Each plot obtained by LBP restoration is averaged over 10000 realizations of the stochastically generated degraded image. 
We can observe that the results obtained by our method in equation (\ref{eq:av-MSE_app}) are in good agreement with those obtained by LBP restoration.

Next, we show the results obtained by setting the function $\xi(S_i, S_j)$ as $\xi(S_i, S_j) = -|S_i - S_j|^2$.
\begin{figure*}[htb]
\begin{center}
\includegraphics[height=4.2cm]{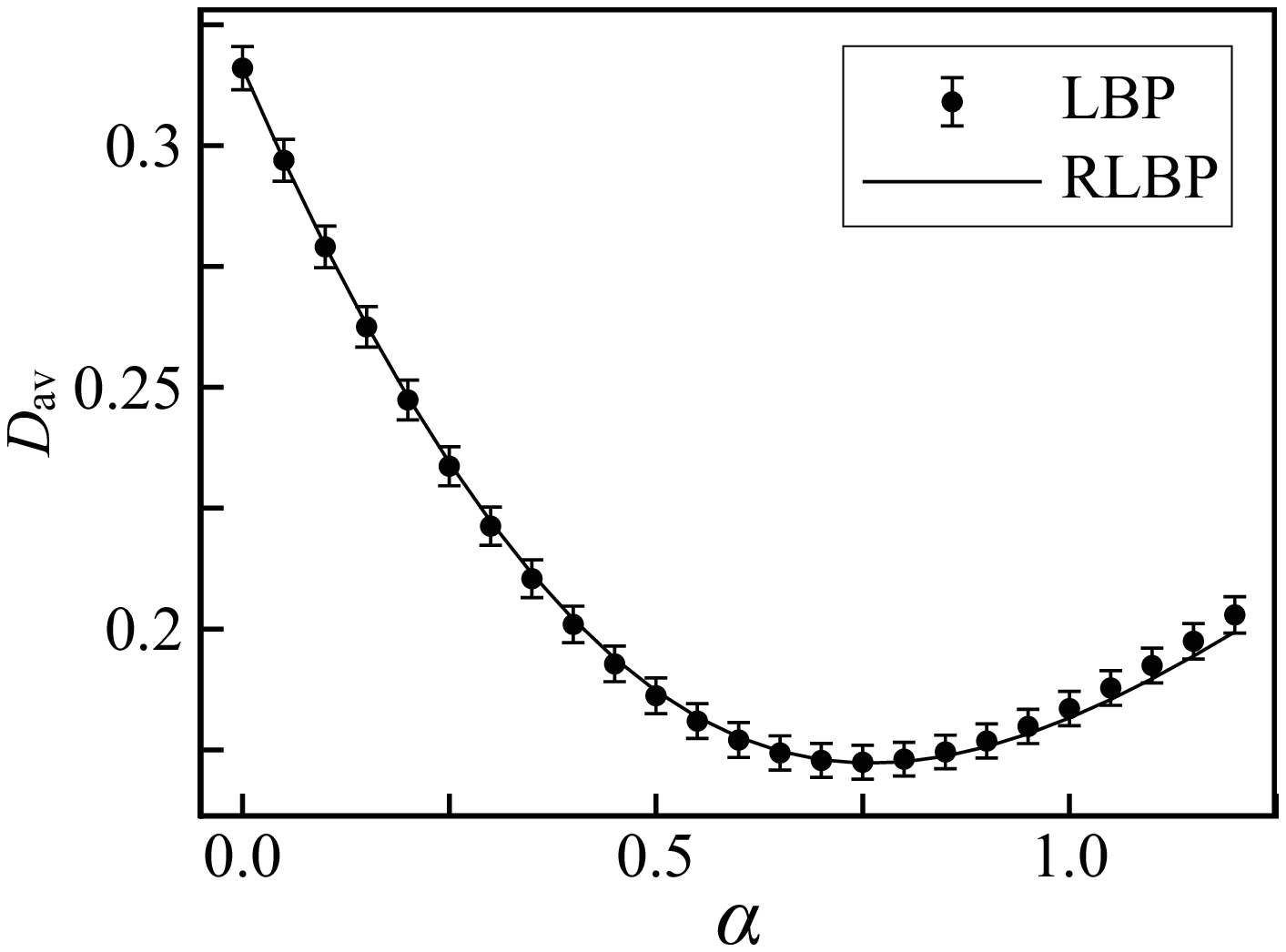}
\includegraphics[height=4.2cm]{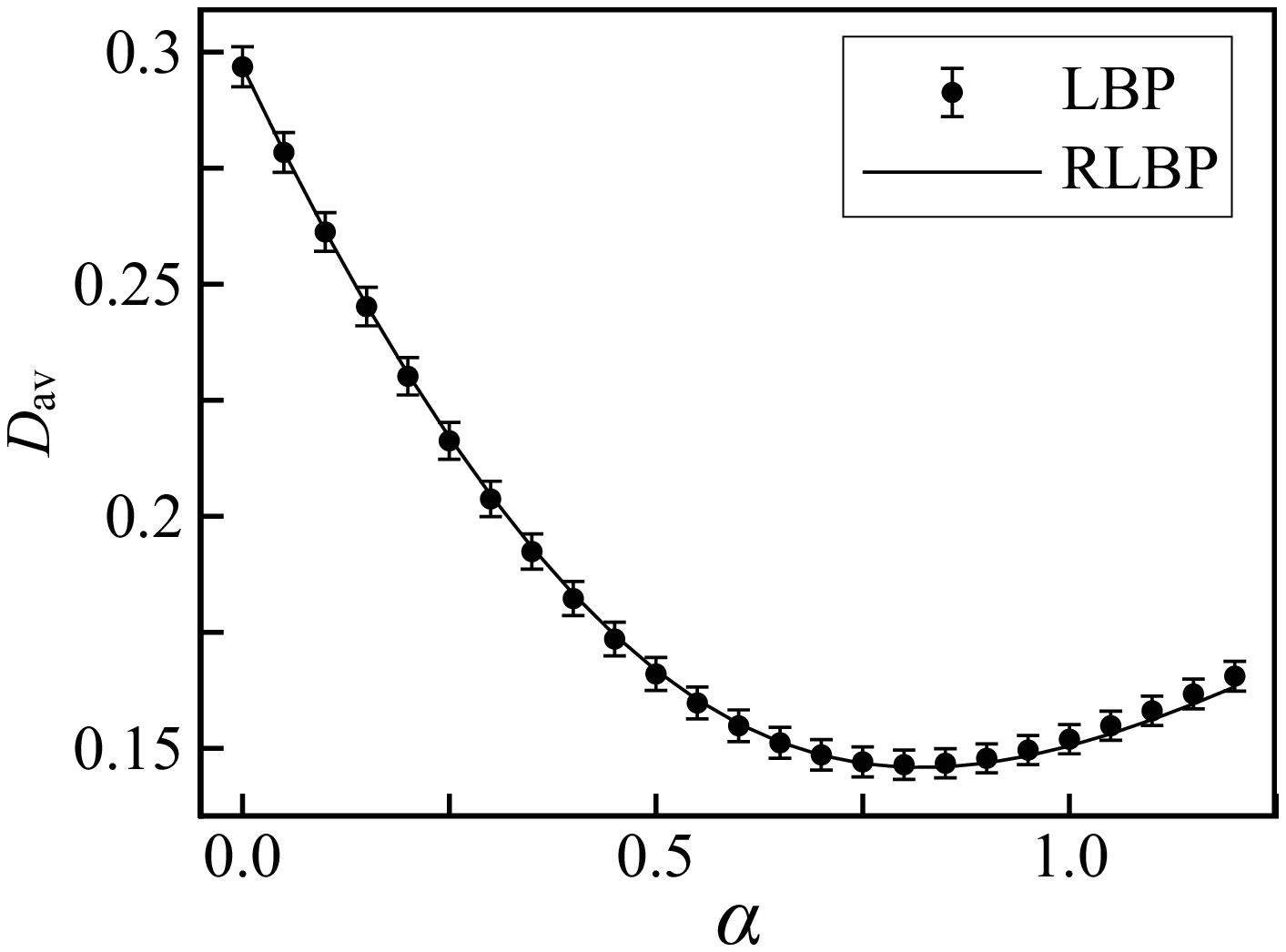}
\includegraphics[height=4.2cm]{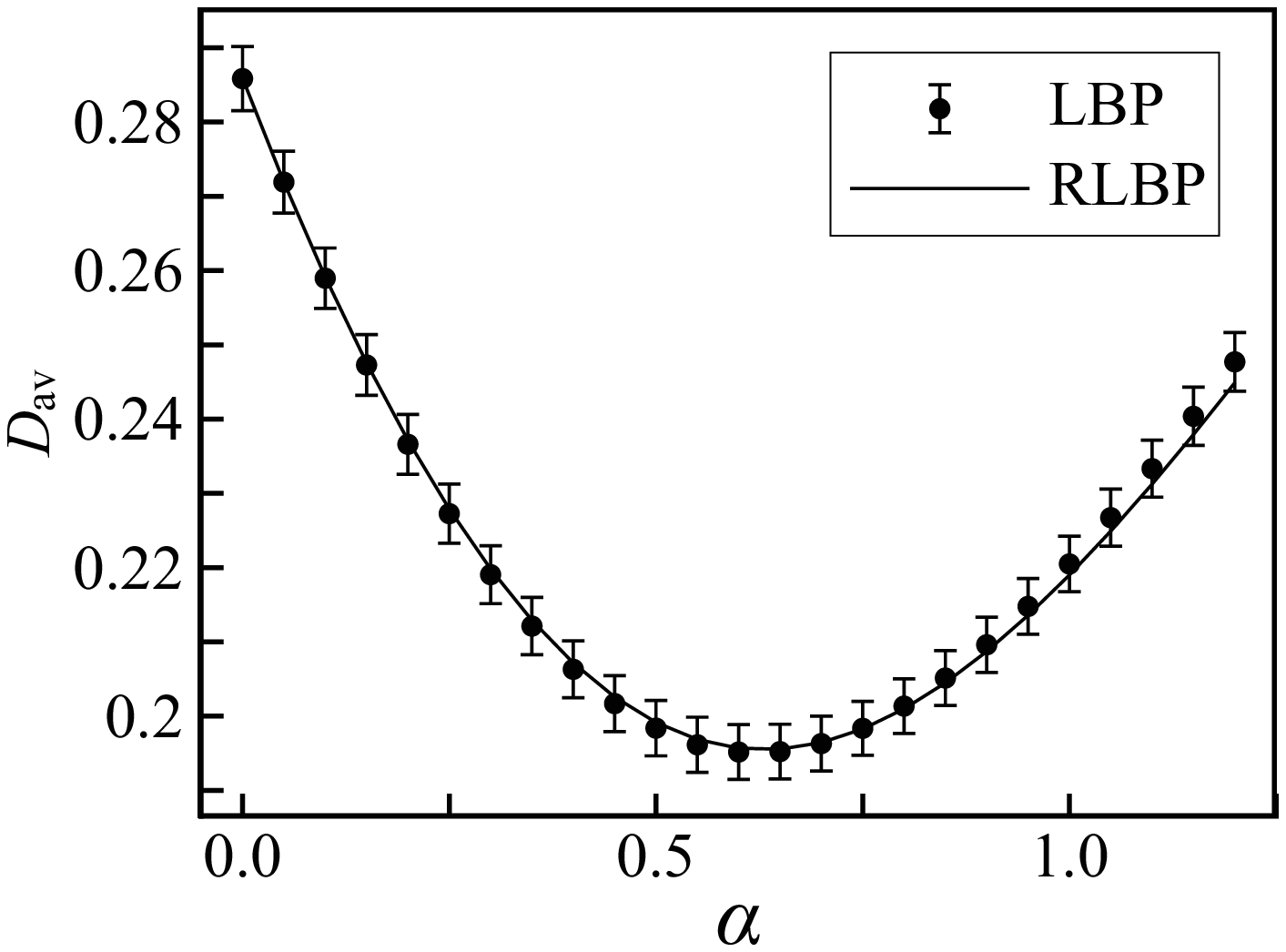}
\end{center}
\caption{Plots of $D_{\mrm{av}}(\bm{I}, \alpha , \sigma^2)$ versus $\alpha$ when $\sigma = 0.5$ and $\xi(S_i, S_j) = -|S_i - S_j|$ 
for the original images shown in figure \ref{fig:original}: the left, middle, and right plots show the results for figures \ref{fig:original}(a), \ref{fig:original}(b), and \ref{fig:original}(c), respectively. 
The error bars are the standard deviation.}
\label{fig:result_al-L1}
\end{figure*}
\begin{figure*}[htb]
\begin{center}
\includegraphics[height=4.2cm]{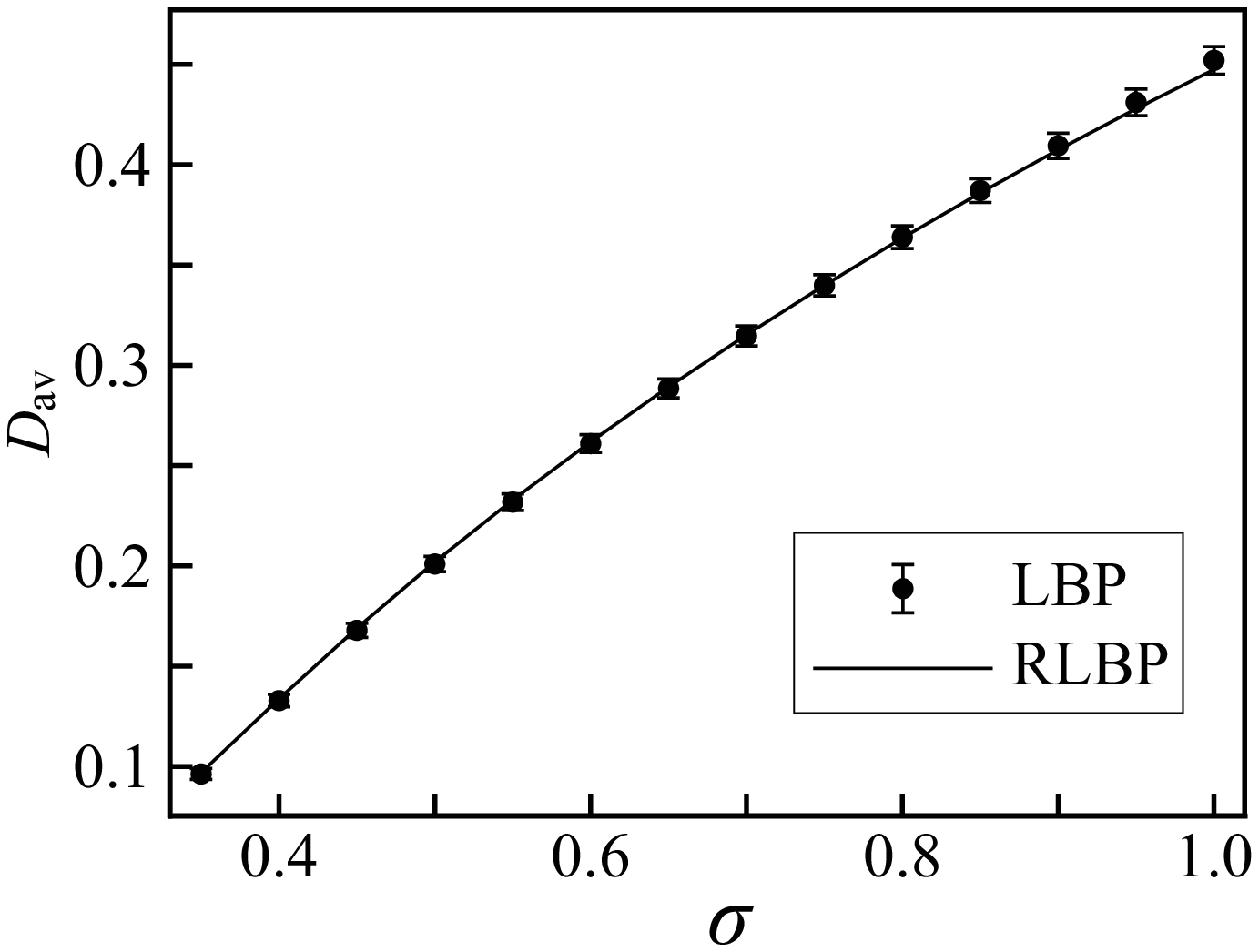}
\includegraphics[height=4.2cm]{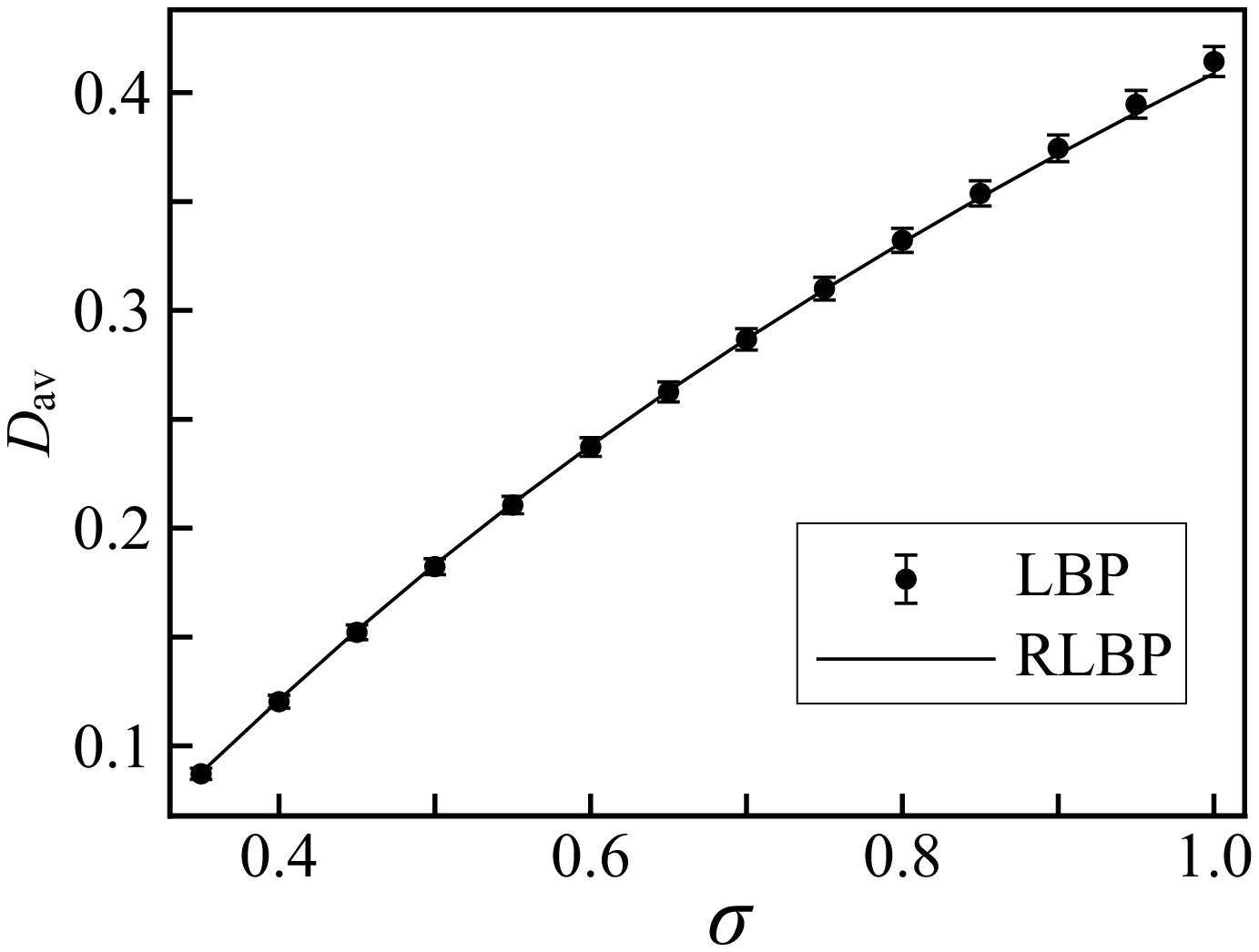}
\includegraphics[height=4.2cm]{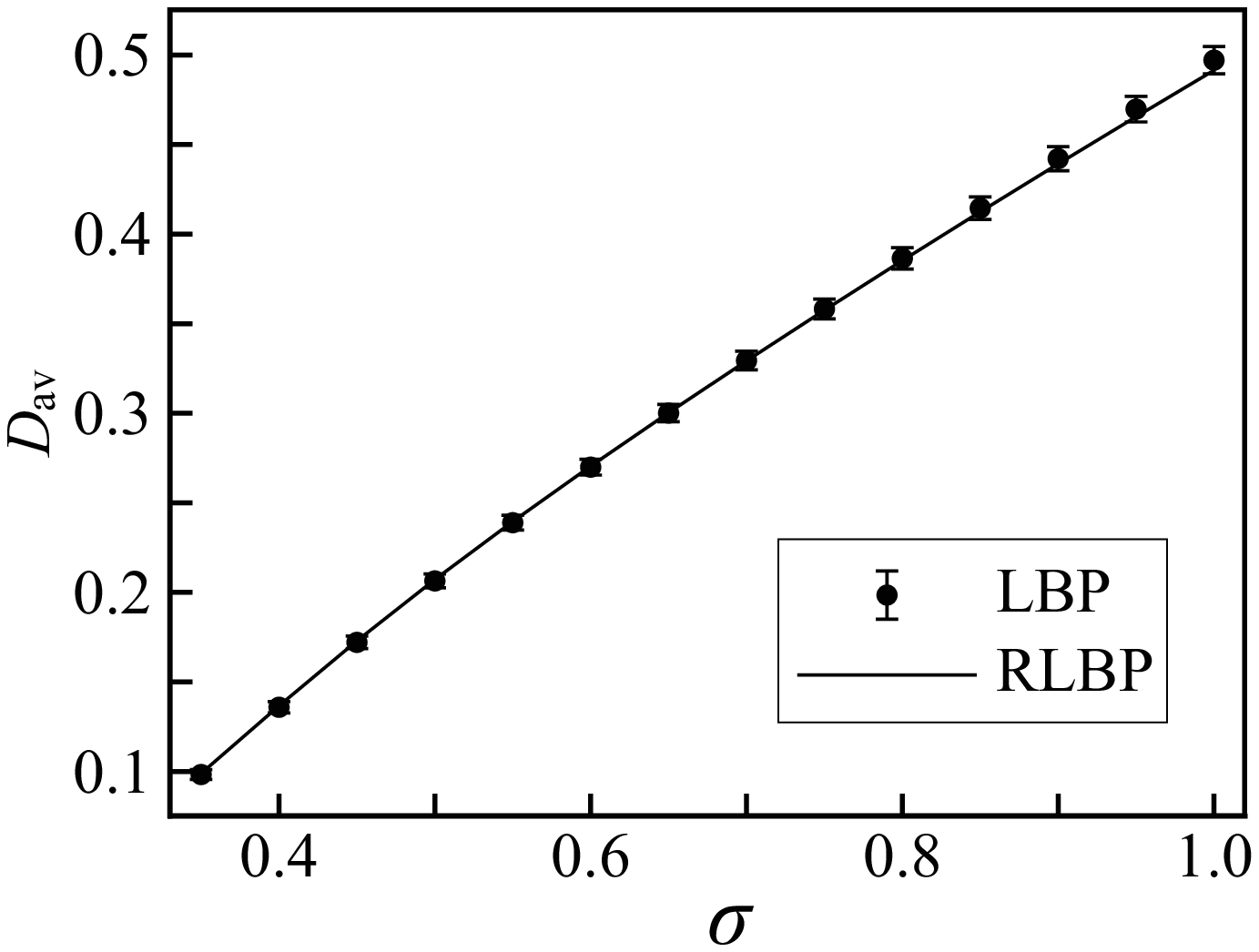}
\end{center}
\caption{Plots of $D_{\mrm{av}}(\bm{I}, \alpha , \sigma^2)$ versus $\sigma$ when $\alpha = 0.4$ and $\xi(S_i, S_j) = -|S_i - S_j|$ 
for the original images shown in figure \ref{fig:original}: the left, middle, and right plots show the results for figure \ref{fig:original}(a), \ref{fig:original}(b), and \ref{fig:original}(c), respectively. 
The error bars are the standard deviation.}
\label{fig:result_sig-L1}
\end{figure*}
Figure \ref{fig:result_al-L1} shows the plots of $D_{\mrm{av}}(\bm{I}, \alpha , \sigma^2)$ versus $\alpha$ when $\sigma = 0.5$, 
and figure \ref{fig:result_sig-L1} shows the plots of $D_{\mrm{av}}(\bm{I}, \alpha , \sigma^2)$ versus $\sigma$ when $\alpha = 0.4$.
The results obtained by our method are in good agreement with those obtained by LBP restoration in figures \ref{fig:result_al} and \ref{fig:result_sig}.

\section{Conclusion and Remarks} \label{sec:conclusion}

In this paper, we proposed an analytical method based on the idea of the RCVM 
to approximately evaluate the quenched average of the Bethe free energy, obtained by LBP, of the pair-wise MRF in equation (\ref{eq:MRF}) in random fields $\bm{h}$. 
Since our message-passing-type formulation allows any form of the functions $\phi_i(S_i, h_i)$ and $\psi_{i,j}(S_i, S_j)$ in the Hamiltonian 
and allows any form of the distributions of the random fields $\{p_i(h_i)\}$, except for the cases where correlations among fields exist, 
the proposed method is applicable to a wide range of applications in physics and in computer science. 
In the argument in section \ref{sec:exact-case}, we found that this approximation is justified in the ferromagnetic mean-field model in random fields.

Although the results obtained by our analytical method in almost all cases were consistent with those obtained by the numerical method, as seen  
in the artificial model presented in section \ref{sec:check-validity} and in the Bayesian image restoration presented in section \ref{sec:Numerical-Bayes}, 
some mismatches, especially in the behaviors of the phase transitions of the magnetizations in section \ref{sec:check-validity}, 
between the results obtained by our method and by the numerical method were observed. 
As mentioned in section \ref{sec:RRG}, the mismatches are considered to be mainly caused by the approximation based on the CVM in equation (\ref{eq:CVM}). 
However, a detailed understanding of its mathematical meaning is still unclear, even though it is one of the most important components of our method, 
because the approximation is for the system replicated by the replica method; 
thus, the development of a mathematical understanding of how it affects the present method is not straightforward.
It should be considered in future studies.

We employed the Bethe approximation in equation (\ref{eq:BetheApproximation-V^alpha}) for the purpose of evaluating the quenched average of the Bethe free energy. 
Our approximate framework, however, allows us to employ other mean-field approximations instead of the Bethe approximation, 
and the resulting form can be expected to be an approximation of an employed approximation method.
Thus, in accordance with the presented framework, it is expected that we can evaluate the quenched averages of the employed mean-field methods for random fields.

\appendix

\section{Derivation of Equation (\ref{eq:quenched-FreeEnergy-Bethe(RS)})}
\label{app:sec:Deriv-BetheFreeEnergy-RS}

By substituting equation (\ref{eq:BetheFreeEnergy-V^alpha}) into equation (\ref{eq:VariationalFreeEnergy-reduce}), we obtain the Bethe approximation of $\mcal{F}_x^{\mrm{RCVM}}[\{Q^{\alpha}, Q_i^{\alpha}\}]$ as
\begin{align}
&\mcal{F}_x^{\mrm{LBP}}[\{Q_i^{\alpha}, Q_{i,j}^{\alpha}\}]
:= \sum_{i \in V}\extr{\bm{\Lambda}_i}\Big\{\sum_{\alpha = 1}^x\sum_{S_i^{\alpha}}\Lambda_i^{\alpha}(S_i^{\alpha})Q_i^{\alpha}(S_i^{\alpha})\nn
&-\frac{1}{\beta}\ln \int \diff h\, p_i(h)\prod_{\alpha = 1}^x\sum_{S_i^{\alpha}} \exp \beta \big( \phi_i(S_i^{\alpha},h) 
+ \Lambda_i^{\alpha}(S_i^{\alpha})\big)\Big\} \nn
&+ \sum_{\alpha = 1}^x \mcal{V}_{\mrm{int}}^{\mrm{bethe}}[\{Q_i^{\alpha},Q_{i,j}^{\alpha}\}]-\frac{1}{\beta}\sum_{i \in V}\sum_{\alpha=1}^x \mcal{H}_1[Q_{i}^{\alpha}].
\label{eq:VariationalFreeEnergy-Bethe}
\end{align}

At the minimum point of this variational free energy, 
we make the RS assumption; that is, 
the relations $Q_i^{\alpha}(S_i) = Q_i(S_i)$ and $Q_{i,j}^{\alpha}(S_i, S_j) = Q_{i,j}(S_i, S_j)$ hold for any $\alpha$. 
Under this assumption, the minimum of $\mcal{F}_x^{\mrm{LBP}}[\{Q_i^{\alpha}, Q_{i,j}^{\alpha}\}]$ is equivalent to 
the minimum of the RS variational free energy expressed as
\begin{align}
&\mcal{F}_x^{\mrm{LBP(RS)}}[\{Q_i, Q_{i,j}\}]
 := \sum_{i \in V}\extr{\bm{\Lambda}_i}\Big\{\sum_{\alpha = 1}^x\sum_{S_i^{\alpha}}\Lambda_i^{\alpha}(S_i^{\alpha})Q_i(S_i^{\alpha})\nn
&-\frac{1}{\beta}\ln \int \diff h\, p_i(h)\prod_{\alpha = 1}^x\sum_{S_i^{\alpha}} \exp \beta \big( \phi_i(S_i^{\alpha},h) 
+ \Lambda_i^{\alpha}(S_i^{\alpha})\big)\Big\} \nn
&+ x \mcal{V}_{\mrm{int}}^{\mrm{bethe}}[\{Q_i,Q_{i,j}\}]-\frac{x}{\beta}\sum_{i \in V} \mcal{H}_1[Q_{i}].
\label{eq:VariationalFreeEnergy-Bethe(RS)}
\end{align}
From the convexity of the first term of this equation with respect to $\bm{\Lambda}_i$ and the extremal conditions of $\bm{\Lambda}_i$, 
it can be ensured that the relation $\Lambda_i^{\alpha}(S_i) =  \Lambda_i(S_i)$ holds for any $\alpha$ at the extremal points of $\bm{\Lambda}_i$. 
Therefore, equation (\ref{eq:VariationalFreeEnergy-Bethe(RS)}) can be reduced to
\begin{align}
&\mcal{F}_x^{\mrm{LBP(RS)}}[\{Q_i, Q_{i,j}\}]
 = \sum_{i \in V}\extr{\Lambda_i}\Big\{x \sum_{S_i}\Lambda_i(S_i)Q_i(S_i)\nn
&-\frac{1}{\beta}\ln \int \diff h\, p_i(h)\exp\Big(x \ln \sum_{S_i} \exp \beta \big( \phi_i(S_i,h) 
+ \Lambda_i(S_i)\big)\Big)\Big\} \nn
&+ x \mcal{V}_{\mrm{int}}^{\mrm{bethe}}[\{Q_i,Q_{i,j}\}]-\frac{x}{\beta}\sum_{i \in V}\mcal{H}_1[Q_{i}],
\label{eq:VariationalFreeEnergy-Bethe(RS)-reduce}
\end{align}
and we regard the minimum of this variational free energy as the Bethe approximation of the true $x$-replicated free energy in equation (\ref{eq:x-replicatedFreeEnergy}). 

From equations (\ref{eq:replica-method}), (\ref{eq:x-replicatedFreeEnergy}), and (\ref{eq:VariationalFreeEnergy-Bethe(RS)-reduce}), 
our desired variational free energy is obtained by
\begin{align}
&\mcal{F}^{\mrm{LBP(RS)}}[\{Q_i, Q_{i,j}\}]\nn
&:=-\frac{1}{\beta}\lim_{x \to 0}\frac{\exp(- \beta \mcal{F}_x^{\mrm{LBP(RS)}}[\{Q_i, Q_{i,j}\}])-1}{x}.
\end{align}
This leads to equation (\ref{eq:quenched-FreeEnergy-Bethe(RS)}).

\section{Derivation of the Message-passing Equation}
\label{app:sec:message-passing}

To perform the conditional minimization of the variational free energy in equation (\ref{eq:quenched-FreeEnergy-Bethe(RS)}), 
we use the Lagrange multipliers as
\begin{widetext}
\begin{align*}
\mcal{L}^{\mrm{LBP(RS)}}[\{Q_i, Q_{i,j}\}]&:=\mcal{F}^{\mrm{LBP(RS)}}[\{Q_i, Q_{i,j}\}]
-\sum_{i \in V}a_i \Big(\sum_{S_i}Q_i(S_i) - 1\Big)-\sum_{\{i,j\} \in E}b_{i,j}\Big(\sum_{S_i,S_j}Q_{i,j}(S_i,S_j) - 1\Big)\nn
\aldef
-\sum_{\{i,j\} \in E}\Big\{\sum_{S_i}\lambda_{j,i}(S_i)\Big(\sum_{S_j}Q_{i,j}(S_i,S_j) - Q_i(S_i)\Big)
+\sum_{S_j}\lambda_{i,j}(S_j)\Big(\sum_{S_i}Q_{i,j}(S_i,S_j) - Q_j(S_j)\Big)\Big\}, 
\end{align*}
\end{widetext}
where the Lagrange multipliers, $\{a_i , b_{i,j}\}$ and $\{\lambda_{i,j}(S_j)\}$, correspond to the normalization constraints in equation (\ref{eq:normalization-constraints-Bethe(RS)}) and 
the marginal constraints in equations (\ref{eq:marginal-constraints-Bethe(RS)-i}) and (\ref{eq:marginal-constraints-Bethe(RS)-j}), respectively.  
From the minimum conditions of $\mcal{L}^{\mrm{LBP(RS)}}[\{Q_i, Q_{i,j}\}]$ with respect to $Q_i(S_i)$ and $Q_{i,j}(S_i,S_j)$, we obtain 
\begin{align}
Q_i(S_i) \propto \exp \frac{\beta}{|\partial i|} \Big( \Lambda_i(S_i)+ \sum_{k \in \partial i}\lambda_{k,i}(S_i)\Big)
\label{eq:MinimumCondition-Qi}
\end{align}
and
\begin{align}
Q_{i,j}(S_i,S_j) \propto \exp \beta \big(\psi_{i,j}(S_i,S_j)+ \lambda_{i,j}(S_j) + \lambda_{j,i}(S_i)\big),
\label{eq:MinimumCondition-Qij}
\end{align}
respectively, where the notation $|\cdots|$ denotes the number of entries of the assigned set.
By introducing the messages in the form
\begin{align*}
\mu_{j \to i}(S_i)&:=\exp \Big\{\frac{\beta}{|\partial i| -1}\Big(\Lambda_i(S_i) - \beta^{-1}\ln Q_i(S_i) \nn
\aldef+ \sum_{k \in \partial i}\lambda_{k,i}(S_i)\Big) - \beta \lambda_{j,i}(S_i)\Big\},
\end{align*}
we obtain
\begin{align}
\exp \big(\beta \lambda_{j,i}(S_i)\big)&=Q_i(S_i)\exp\big(- \beta\Lambda_i(S_i) \big)\nn
\aleq
\times \prod_{k \in \partial i \setminus \{j\}}\mu_{k \to i}(S_i).
\label{eq:exp(gamma)}
\end{align}
From equations (\ref{eq:MinimumCondition-Qi}) and (\ref{eq:exp(gamma)}), we obtain the relation 
\begin{align}
\beta \Lambda_i(S_i) = \sum_{k \in \partial i } \ln \mu_{k \to i}(S_i) + c_i,
\label{eq:exp(Lambda)-appendix}
\end{align}
where $c_i$ is a constant unrelated to $S_i$. 
Since the value of $c_i$ does not affect our results, without loss of generality, we set $\{c_i\}$ to zeros.   

By substituting equations (\ref{eq:MinimumCondition-Qi}) and (\ref{eq:MinimumCondition-Qij}) 
into the marginal constraints in equations (\ref{eq:marginal-constraints-Bethe(RS)-i}) and (\ref{eq:marginal-constraints-Bethe(RS)-j}) 
and using equation (\ref{eq:exp(gamma)}), 
we obtain the message-passing equation as
\begin{align}
\mu_{j \to i}(S_i) &\propto\sum_{S_j}Q_j(S_j)\exp\beta \big(-\Lambda_j(S_j) + \psi_{i,j}(S_i,S_j)\big)\nn
\aleq 
\times\prod_{k \in \partial j \setminus \{i\}}\mu_{k \to j}(S_j)\nn
&\propto \sum_{S_j}Q_j(S_j)\exp \big( \beta\psi_{i,j}(S_i,S_j)\big)\mu_{i \to j}(S_j)^{-1}.
\label{eq:MessagePassing-LBP(RS)-appendix}
\end{align}
Here, from the first line to the second line of this equation, we use the relation in equation (\ref{eq:exp(Lambda)-appendix}).
From the extremal conditions for $\{\Lambda_i(S_i)\}$ in the first term in equation (\ref{eq:quenched-FreeEnergy-Bethe(RS)}), we obtain equation (\ref{eq:Qi-LBP(RS)}).
From equations (\ref{eq:MinimumCondition-Qij}), (\ref{eq:exp(gamma)}), and (\ref{eq:exp(Lambda)-appendix}), the two-vertex marginal distributions $\{Q_{i,j}(S_i,S_j)\}$ can be expressed as
equation (\ref{eq:Qij-LBP(RS)}).

\section{Approximation in Equation (\ref{eq:av-MSE_app})}
\label{app:sec:Deriv-D_av}

Using the belief obtained by LBP and any real value $\gamma$, we define the new distribution as
\begin{align}
&\zeta_i(S_i\mid \gamma):=\frac{b_i^{\gamma}(S_i)}{\sum_{S_i}b_i^{\gamma}(S_i)}\nn
&=\frac{\exp \gamma\big( \beta \phi_i(S_i,h_i)  +\sum_{j \in \partial i}\ln M_{j \to i}(S_i) \big)}
{\sum_{S_i}\exp \gamma\big( \beta \phi_i(S_i,h_i)  +\sum_{j \in \partial i}\ln M_{j \to i}(S_i) \big)}.
\label{eq:Def-Bi}
\end{align} 
Assuming $b_i^{\gamma}(S_i)$ has a unique maximum with respect to $S_i$, we obtain the equality
\begin{align}
\Big[ \big(\argmax_{S_i} b_i(S_i)\big)^k\Big]_{\bm{h}}=\lim_{\gamma \to \infty}\sum_{S_i}S_i^k \big[ \zeta_i(S_i \mid \gamma)\big]_{\bm{h}},
\label{eq:transform-maxBi}
\end{align}
for any $k$.

As mentioned in section \ref{sec:ReplicaMessagePassing}, the distribution $Q_i(S_i)$ in equation (\ref{eq:Qi-LBP(RS)}),
\begin{align*}
Q_i(S_i) = \int \diff h\, p_i(h)q_i(S_i\mid  h),
\end{align*}
where
\begin{align*}
q_i(S_i \mid h):=\frac{\exp \beta \big( \phi_i(S_i,h) + \Lambda_i(S_i)\big)}{ \sum_{S_i} \exp \beta \big( \phi_i(S_i,h) + \Lambda_i(S_i)\big)},
\end{align*}
is regarded as the approximation of the quenched average of the belief, $[ b(S_i)]_{\bm{h}}$.
According to equation (\ref{eq:Def-Bi}), we define the distribution as
\begin{align*}
&\rho_i(S_i \mid h, \gamma) := \frac{q_i(S_i\mid  h)^{\gamma}}{\sum_{S_i} q_i(S_i\mid  h)^{\gamma}}\nn
&=\frac{\exp \gamma\beta \big( \phi_i(S_i,h) + \Lambda_i(S_i)\big)}{ \sum_{S_i} \exp \gamma\beta \big( \phi_i(S_i,h) + \Lambda_i(S_i)\big)}.
\end{align*}
As mentioned in section \ref{sec:ReplicaMessagePassing}, 
when the distributions of the random fields are Dirac delta functions, $p_i(h) = \delta(h - h_i)$, 
the proposed method is reduced to standard LBP.
Thus, in this case, since $q_i(S_i \mid h_i) = b_i(S_i)$, the equality $\rho_i(S_i \mid h_i, \gamma) = \zeta_i(S_i\mid \gamma)$ holds. 
From this relation, it is expected that the quenched average of $\zeta_i(S_i\mid \gamma)$ is approximated as
\begin{align*}
\big[ \eta_i(S_i \mid \gamma)\big]_{\bm{h}}\approx \int \diff h\, p_i(h)\rho_i(S_i \mid h, \gamma).
\end{align*}
Using this approximation, we approximate the right-hand side of equation (\ref{eq:transform-maxBi}) by
\begin{align}
&\lim_{\gamma \to \infty}\sum_{S_i}S_i^k \big[ \zeta_i(S_i \mid \gamma)\big]_{\bm{h}}\nn
&\approx \lim_{\gamma \to \infty}\sum_{S_i}S_i^k \int \diff h\, p_i(h)\rho_i(S_i \mid h, \gamma)\nn
&=\int \diff h\, p_i(h)r_i(h)^k,
\label{eq:maxBi_approx}
\end{align}
where
\begin{align*}
r_i(h):=\argmax_{S}\big( \phi_i(S,h) + \Lambda_i(S) \big).
\end{align*}

Using equations (\ref{eq:MPM-LBP}), (\ref{eq:transform-maxBi}), and (\ref{eq:maxBi_approx}), we obtain the approximation as
\begin{align}
\big[\big(S_i^{\mrm{MPM(LBP)}}\big)^k \big]_{\bm{h}} \approx \int \diff h\, p_i(h)r_i(h)^k,
\end{align}
for any $k$, where $p_i(h) = \mcal{N}(h \mid I_i, \sigma^2)$.
By using this approximation,
\begin{align*}
D_{\mrm{av}}(\bm{I}, \alpha , \sigma^2) &= \frac{1}{n}\sum_{i \in V} \Big(I_i^2 -2 I_i\big[S_i^{\mrm{MPM(LBP)}} \big]_{\bm{h}}  \nn
\aleq
- \big[\big(S_i^{\mrm{MPM(LBP)}}\big)^2 \big]_{\bm{h}}\Big) 
\end{align*}
is approximated by
\begin{align*}
D_{\mrm{av}}(\bm{I}, \alpha , \sigma^2) &\approx \frac{1}{n}\sum_{i \in V} \Big(I_i^2 -2 I_i\int \diff h\, p_i(h)r_i(h)  \nn
\aleq
- \int \diff h\, p_i(h)r_i(h)^2\Big) \nn
&=\frac{1}{n}\sum_{i \in V}\int \diff h\,p_i(h)\big(I_i - r_i(h)\big)^2.
\end{align*}

\subsection*{Acknowledgment}
This work was supported by CREST, JST, and a Grant-In-Aid (Nos. 24700220 and 25280089) 
for Scientific Research from the Ministry of Education, Culture, Sports, Science and Technology, Japan.
\appendix

\bibliography{citations}
\end{document}